\documentclass[preprint,12pt,nopreprintline]{elsarticle}

\usepackage{amssymb}
\usepackage{lscape}
\usepackage{amsmath}
\usepackage{svg}
\usepackage{url}

\journal{Knowledge-Based Systems}

\begin{document}

\begin{frontmatter}



\title{CrisisTransformers: Pre-trained language models and sentence encoders for crisis-related social media texts}


\author[inst1]{Rabindra Lamsal\corref{cor1}}

\affiliation[inst1]{organization={School of Computing and Information Systems},
            addressline={The University of Melbourne}, 
            country={Australia}}

\author[inst1]{Maria Rodriguez Read}
\author[inst1]{Shanika Karunasekera}

\cortext[cor1]{rlamsal@student.unimelb.edu.au}

\begin{abstract}
Social media platforms play an essential role in crisis communication, but analyzing crisis-related social media texts is challenging due to their informal nature. Transformer-based pre-trained models like BERT and RoBERTa have shown success in various NLP tasks, but they are not tailored for crisis-related texts. Furthermore, general-purpose sentence encoders are used to generate sentence embeddings, regardless of the textual complexities in crisis-related texts. Advances in applications like text classification, semantic search, and clustering contribute to the effective processing of crisis-related texts, which is essential for emergency responders to gain a comprehensive view of a crisis event, whether historical or real-time. To address these gaps in crisis informatics literature, this study introduces \textit{CrisisTransformers}, an ensemble of pre-trained language models and sentence encoders trained on an extensive corpus of over 15 billion word tokens from tweets associated with more than 30 crisis events, including disease outbreaks, natural disasters, conflicts, and other critical incidents. We evaluate existing models and CrisisTransformers on 18 crisis-specific public datasets. Our pre-trained models outperform strong baselines across all datasets in classification tasks, and our best-performing sentence encoder improves the state-of-the-art by 17.43\% in sentence encoding tasks. Additionally, we investigate the impact of model initialization on convergence and evaluate the significance of domain-specific models in generating semantically meaningful sentence embeddings. The models are publicly available at: \texttt{\url{https://huggingface.co/crisistransformers}}
\end{abstract}




\end{frontmatter}


\section{Introduction}
Social media platforms, such as Facebook and Twitter, have become an essential medium for information sharing and communication during times of crisis \cite{imran2015processing,lamsal2022socially}. Particularly during disasters, such as wildfires, earthquakes, hurricanes, tsunamis, floods, cyclones, and epidemics, social media platforms play a pivotal role in the timely dissemination of information \cite{starbird2010pass,thomson2012trusting,alam2018twitter,pourebrahim2019understanding,lamsal2023narrative}. These platforms are critical information sources for affected individuals and emergency responders, enabling real-time updates on evolving situations \cite{sarcevic2012beacons,hughes2009twitter,vieweg2010microblogging,vieweg2012situational} and providing firsthand accounts from those directly and indirectly impacted \cite{imran2015processing,lamsal2022socially}. The enormous amount of user-generated content on social media platforms acts as a rich source of historical as well as real-time data. However, the volume \cite{stieglitz2018social} and textual complexity of crisis-related social media texts give rise to multiple challenges for effective analysis and understanding. The volume necessitates automated analysis as the number of conversations increases exponentially during a crisis, and the textual complexity involves dealing with informally written texts with a significant presence of acronyms, misspellings, hashtags, mentions, etc.

Transformer-based \cite{vaswani2017attention} domain-specific pre-trained language models have helped produce state-of-the-art results for numerous NLP tasks in various areas such as biomedical research \cite{lee2020biobert}, scientific literature analysis \cite{beltagy-etal-2019-scibert}, clinical text analysis \cite{huang2019clinicalbert} and financial text analysis \cite{yang2020finbert}. Trained on massive amounts of domain-specific texts, these models produce contextual text representations within their respective domains. Despite the broad array of domains in which pre-trained models have been employed, a notable gap exists, i.e., the absence of pre-trained models explicitly tailored for crisis-related social media texts. Furthermore, pre-trained language models do not produce semantically rich sentence embeddings\footnote{Semantically rich sentence embeddings position semantically similar sentences close together in the vector space.}, critical for tasks like semantic search and clustering \cite{reimers2019sentence}. Currently, the generation of semantically meaningful sentence embeddings, regardless of the domain, relies on general-purpose sentence embedding models (sentence encoders) \cite{reimers2019sentence, gao2021simcse}. These models utilize pre-trained models that have been trained on corpora comprising texts from broad and general domains. Studies have consistently found BERT \cite{devlin-etal-2019-bert} and its variants to be effective in various crisis-related tasks, including informativeness and humanitarian classification \cite{alam2021crisisbench}, health and figurative mentions identification \cite{biddle2020leveraging}, emotion classification \cite{myint2024unveiling}, COVID-19 data analysis \cite{klein2021toward,lamsal2023narrative}, and more. Hence, there exists a necessity to investigate the efficacy of utilizing domain-specific pre-trained language models and sentence encoders for processing crisis-related social media texts.

To address the above-discussed gaps in the crisis informatics literature, this study proposes \textit{CrisisTransformers}, an ensemble of pre-trained language models and sentence encoders trained on hundreds of millions of crisis-related tweets from over 30 different crisis events, including the COVID-19 pandemic. CrisisTransformers' embeddings can be utilized in various tasks, including text classification \cite{ashktorab2014tweedr,caragea2011classifying,imran2013extracting,li2018comparison}, semantic search \cite{dutt2019utilizing}, clustering \cite{ashktorab2014tweedr,curiskis2020evaluation,lamsal2022socially}, and topic modelling \cite{grootendorst2022bertopic}. Advancements in these applications contribute to a more comprehensive understanding of crisis-related social media texts, thereby aiding decision-making processes and facilitating targeted interventions and communication strategies during times of crisis \cite{imran2015processing}.

This study contributes the following to the \textbf{crisis informatics} literature:

\begin{itemize}
    \item We provide the first set of experiments relative to domain-specific pre-training to address the following research questions:

        \begin{itemize}
            \item How does the choice of model initialization impact pre-training in terms of loss convergence?
            \item With BERTweet \cite{nguyen2020bertweet} and other strong baselines in place, can yet another domain-specific pre-trained model demonstrate superior performance in crisis-related social media text classification?
            \item To what extent do domain-specific pre-trained models help generate sentence embeddings with semantic richness, in comparison to current pre-trained models and sentence encoders?
        \end{itemize}
    \item We introduce CrisisTransformers, the first pre-trained language models and sentence encoders designed for processing crisis-related social media texts. The pre-training of CrisisTransformers was done on 6 NVIDIA A100 GPUs over a period of 2 months.
    \item Our pre-trained models outperform existing models across all 18 crisis-related datasets in classification tasks, and our best-performing sentence encoder improves the current state-of-the-art by 17.43\% in sentence encoding tasks. Results confirm that CrisisTransformers can capture distinct linguistic nuances, informal language structures, and unique contextual cues present in crisis contexts.
    \item We publicly release CrisisTransformers. The released models can be used with the \textit{Transformers} \cite{wolf2019huggingface} library. We anticipate that these models will serve as a robust baseline for tasks involving the analysis of crisis-related social media texts.
\end{itemize}

The rest of the paper is organized as follows: Section \ref{relatedwork} discusses related work, Section \ref{methods} details the materials and methods used in designing CrisisTransformers, Section \ref{results} presents evaluation results and discussions, and Section \ref{conclusion} concludes the paper.

\section{Related Work}
\label{relatedwork}
The current landscape of crisis informatics heavily relies on Transformer-based models (BERT-family \cite{devlin-etal-2019-bert}) trained on general domain texts (we discuss the relevant literature on BERT-family later in this section). For instance, \cite{alam2021crisisbench} conducted classification tasks on 8 human-annotated crisis datasets and reported BERT and RoBERTa as the best classifiers. \cite{biddle2020leveraging} demonstrated that BERT performs better in the correct classification of health and figurative mentions on Twitter. Similarly, the BERT family has also been employed in the identification and classification of transportation disaster tweets \cite{prasad2023identification}, classifying informative tweets \cite{koshy2023multimodal}, identifying location mentions on disaster tweets \cite{suwaileh2023idrisi}, and emotion classification in crisis-related tweets \cite{myint2024unveiling}. Likewise, the BERT family has also been extensively utilized in COVID-19 data to create state-of-the-art classifiers for vaccine-related stance \cite{poddar2022winds,cotfas2021longest,hayawi2022anti}, inferring the origin locations of tweets \cite{lamsal2022did}, distinguishing tweets that self-report potential cases \cite{klein2021toward}. Furthermore, the state-of-the-art sentence embeddings from Sentence-Transformers are based on the BERT family.

Generally, transformer-based models fall into three main categories: encoder-decoder models, decoder-only models, and encoder-only models. Auto-regressive models, such as GPT-like \cite{brown2020language} models (CTRL, GPT, GPT-2, Transformer XL), utilize only the decoder component of the Transformer architecture. They focus on predicting the subsequent word in a sentence, making them optimal for text generation tasks. In such models, the attention mechanism operates such that it can only access preceding words, making them autoregressive in nature. On the other hand, BART/T5-like \cite{lewis2019bart,raffel2020exploring} models (BART, mBART, Marian, T5), known as sequence-to-sequence models \cite{zhong2023e2s2}, use both the encoder and decoder components of the Transformer architecture. These models are best suited for generating new sentences based on provided input sequences, such as in text summarization, translation, and question answering tasks. In such models, the attention mechanism in the encoder accesses all words in the input sequence, while in the decoder, it can only access preceding words. Lastly, there are BERT-like models \cite{devlin-etal-2019-bert} (BERT, RoBERTa, ALBERT, ELECTRA), which are auto-encoding models that use only the encoder component of the Transformer architecture. These models are ideal for tasks that require the entire input sequence to make decisions, such as text classification and named-entity recognition. In these models, the attention mechanism accesses all words in the input sequence, a feature commonly referred to as bi-directional attention. This characteristic makes encoder models well-suited for tasks requiring contextual embeddings. Next, we review the state-of-the-art encoder-only Transformer models, which are the focus of this study.

BERT \cite{devlin-etal-2019-bert} has become a ubiquitous baseline in NLP tasks. BERT uses two pre-training objectives~---~masked language modelling (MLM) and next sentence prediction (NSP). The MLM objective involves randomly masking specific tokens of an input sentence and training the model to predict the original masked tokens based on the context (surrounding words). Through this objective, BERT learns relationships between words and captures rich contextualized representations. Since the introduction of BERT, MLM has become a standard pre-training objective for many transformer-based models. Various improvements in training approaches and variants of MLM have been explored in subsequent research. RoBERTa, proposed in \cite{liu2019roberta}, outperformed BERT in various downstream tasks with some changes in the pre-training process: large batch size, longer training, more training data, and removal of the NSP objective. In \cite{lan2019albert}, ALBERT was introduced, which offered competitive results with reduced parameters through factorized embedding parameterization and cross-layer parameter sharing. MPNet was introduced in \cite{song2020mpnet} combining MLM and permuted language modelling (PLM). In PLM \cite{yang2019xlnet}, a sequence is randomly permuted, and the model autoregressively predicts the tokens. In \cite{conneau2019unsupervised}, XLM-RoBERTa was trained to confirm the usefulness of pre-training multilingual language models on large-scale data for cross-lingual transfer tasks. ELECTRA, introduced in \cite{clark2020electra}, proposed a pre-training objective where two models (generator and discriminator) are involved~---~the generator replaces tokens in a sequence, and the discriminator predicts which tokens are originals and which are the ones replaced by the generator. The above-discussed models were pre-trained on datasets such as \textit{Wikipedia}, \textit{BooksCorpus}, \textit{OpenWebText}, \textit{CC-News}, etc., which contain general domain texts. Researchers have also introduced domain-specific pre-trained models; we discuss some of those models next.

BERTweet \cite{nguyen2020bertweet} is a transformer-based model specifically designed for processing Twitter data and other social media texts. It leverages the BERT model configuration and incorporates RoBERTa's pre-training approach. During pre-training, it was exposed to a massive corpus containing 16 billion word tokens. BioBERT, which was introduced in \cite{lee2020biobert}, was pre-trained on biomedical texts, including PubMed abstracts (PubMed) and PubMed Central full-text articles (PMC), using the same architecture as BERT. Similarly, SciBERT \cite{beltagy-etal-2019-scibert} also shared the architecture of BERT but was pre-trained on a random sample of over 1 million papers. Its pre-training corpus consisted of 18\% computer science and 82\% biomedical domain full-text papers. Additionally, a variant of the BERT architecture called ClinicalBERT \cite{huang2019clinicalbert} was developed by pre-training on electronic health records. This specific pre-training made ClinicalBERT suitable for processing clinical text and medical data. BERT's application has also been extended to the finance domain. FinBERT, introduced in \cite{yang2020finbert}, is a pre-trained model trained on an extensive financial communication corpus containing 4.9 billion tokens.

When the pre-trained models utilize either the embeddings of the \textit{CLS} token or the mean-pooling of all tokens to generate \textit{sentence embeddings} and subsequently undergo fine-tuning, they produce state-of-the-art results in text classification/regression tasks. However, previous research shows that such sentence embeddings lack semanticity and are actually worse than averaging GloVe embeddings \cite{reimers2019sentence}. For effective semantic search and clustering tasks, it is critical to have semantically meaningful embeddings that position sentences in a vector space, such that semantically similar sentences are located closely together. Generating such sentence embeddings is an extensively researched area, and various methods have been proposed, which we discuss next.

In \cite{kiros2015skip}, an encoder-decoder was trained to reconstruct the surrounding sentences of an encoded sequence so that the sentences that share semantic properties are mapped to similar vector representations. In \cite{conneau2017supervised}, a siamese BiLSTM network was trained with max-pooling on the Stanford Natural Language Inference (SNLI) dataset which outperformed previous unsupervised methods \cite{kiros2015skip,hill2016learning}. In \cite{cer2018universal}, a transformer network was trained and unsupervised learning was extended with training on the SNLI dataset. Additionally, \cite{yang2018learning} presented an unsupervised learning approach to sentence-level semantic similarity based on conversational data. Until this period, the sentence encoding approaches involved training the respective networks from scratch. After the introduction of BERT in 2018, replacing the unsupervised training part of designing sentence encoders became possible. In \cite{reimers2019sentence}, BERT was finetuned through siamese and triplet networks on SNLI and Multi-Genre natural language inference (MultiNLI) datasets, with softmax classifier over ``contradiction", ``entailment", and ``neutral" labels. Similarly, \cite{gao2021simcse} proposed a contrastive approach (SimCSE) to finetune pre-trained models with natural language inference datasets using ``contradiction" pairs as hard negatives. Following \cite{gao2021simcse}, \cite{reimers2019sentence} fine-tuned multiple pre-trained models using the contrastive training objective on over 1 billion sentence pairs and publicly released all their models as Sentence-Transformers.

The substantial computational resources large language models require pose challenges for real-time processing, especially in contexts like analyzing social media posts where rapidity is critical. Models with smaller footprints, typically comprising millions rather than billions of parameters, emerge as promising alternatives \cite{zhong2023can} as they offer viable solutions for scenarios where computational resources are limited. Furthermore, the closed-source nature of some large models, which are accessible only through APIs, introduces obstacles in terms of transparency and adaptability to specific research or application requirements. Smaller models not only address computational constraints but also promote transparency and flexibility in model usage. Therefore, in this study, we consider the \textit{base} architectures of MPNet, BERTweet, BERT, RoBERTa, XLM-RoBERTa, ALBERT, and ELECTRA, as baselines for the classification task, and Sentence-Transformers and SimCSE as baselines for the sentence encoding task. These baselines (except ALBERT) share similar parameter counts with the models proposed in this study.

\section{Materials and methods}
\label{methods}

\begin{figure}
    \centering
    \includegraphics[width=1\textwidth]{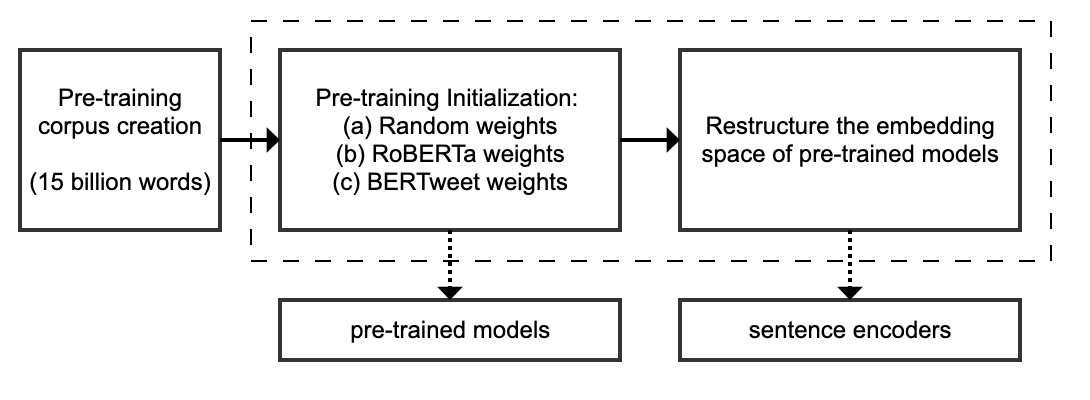}
    \caption{A high-level methodological view for developing pre-trained models and sentence encoders.}
    \label{overview}
\end{figure}

In this section, we detail the curation process of the pre-training corpus (Section \ref{crisis-corpus}) and the development of pre-trained models (Section \ref{unsup-pre-training}) and sentence encoders (Section \ref{sentence-encoders}). A high-level methodological view is preovided in Figure \ref{overview}.

\subsection{The crisis corpus}
\label{crisis-corpus}
\begin{figure}
    \centering
    \includegraphics[width=0.9\textwidth]{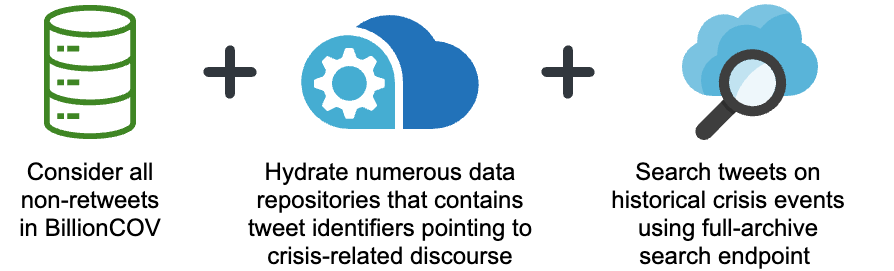}
    \caption{The pre-training corpus curation process.}
    \label{pretraining_corpus}
\end{figure}

A large-scale social media text corpus was curated for unsupervised pre-training, with Twitter serving as the primary data source. Our main objective was to create a comprehensive corpus containing texts discussing a diverse range of crisis events, such as disease outbreaks, natural disasters, terrorist attacks, conflicts, and other critical incidents. In general, as illustrated in Figure \ref{pretraining_corpus}, the corpus underwent curation across three distinct stages: (i) consideration of an in-house dataset consisting of billions of tweets, (ii) hydration of Twitter identifiers collected from various data repositories, and (iii) utilization of Twitter's full-archive endpoint to search historical tweets. We maintained an in-house billion-scale COVID-19 tweets dataset from the onset of the COVID-19 outbreak until March 2023. The initial version of the dataset, \textit{COV19Tweets} \cite{lamsal2021design}, comprised more than 2.2 billion tweets. Subsequently, we created the second version, \textit{BillionCOV} \cite{lamsal2023billioncov}, by filtering out unavailable tweets, resulting in over 1.4 billion tweets. For this study, we considered all the tweets present in BillionCOV, excluding retweets. Although BillionCOV contains COVID-19-related tweets, the COVID-19 discourse was not solely limited to discussions about the virus. Numerous other events unfolded worldwide along with the pandemic, including economic crises, natural disasters, humanitarian crises, social unrest, mental health concerns, and social issues.

Next, we collected tweet identifiers from multiple data repositories such as \textit{CrisisNLP} \cite{imran2016twitter} and \textit{DocNow Catalog}\footnote{https://catalog.docnow.io/}. Tweets collected from these sources needed to be hydrated to re-create the datasets locally, as Twitter's data re-distribution policy restricts sharing data other than tweet identifiers. At this stage, the corpus had texts related to more than 30 crisis events that occurred after 2014. Furthermore, to fill the temporal gap in the corpus, we utilized Twitter's full-archive endpoint to search for historical tweets created between 2006 and 2013. We applied \textit{lang:en} condition and used the following keywords (along with their \#hashtag and plural variants): crisis, disaster, earthquake, typhoon, volcano, flood, landslide, hurricane, tornado, cyclone, wildfire, famine, drought, tsunami, avalanche, epidemic, hailstorm, storm, protest, virus, war, and riot. Below are some of the crisis events covered in the corpus.

\begin{itemize}
    \item Disease Outbreaks: COVID-19, Middle East Respiratory Syndrome, Ebola Virus Outbreak.
    \item Natural Disasters: Hurricanes Harvey, Irma, Florence, Dorian, Odile, Cyclone PAM, Typhoon Hagupit, California Earthquake, Pakistan Earthquake, Chile Earthquake, Nepal Earthquake, Pakistan Floods, India Floods, Iceland Volcano, Tropical Storm Imelda.
    \item Terrorist Attacks: Paris Attacks, Stockholm Attack, Catalonia Attacks, Peshawar School Attack.
    \item Protests and Activism: \#J20, Tyendinaga protests.
    \item Shootings: Dallas Police Shooting, Las Vegas Shooting.
    \item Landslides: Landslides worldwide.
    \item Conflicts: Gaza, Palestine Conflict.
    \item Civil War: Fall of Aleppo.
    \item Missing Flight: Flight MH370.
\end{itemize}

\subsubsection{Text pre-processing}
Each tweet in the corpus was pre-processed as follows: We (i) replaced URLs with ``HTTPURL" token, (ii) replaced mentions (usernames) with ``@USER" token (iii) decoded HTML entities to their original form (e.g., \&amp; to \&), (iv) removed newline characters and replaced multiple consecutive whitespaces with a single space, (v) fixed text encoding to correct various encoding issues and improve consistency in text representation, and (vi) replaced emojis with their textual representation, as their descriptive text counterparts are meaningful. We considered only the tweets with more than ten tokens. Refer to Table \ref{corpus_stats} for the descriptive statistics of the corpus.

\begin{table}
\scriptsize
    \centering
    \begin{tabular}{c|c}
    \hline
         & \textbf{count} \\
         \hline \hline
      tokens   & 15 billion\\
      sentences & 997 million\\
      unique tokens & 36.7 million\\
      \hline
    \end{tabular}
    \caption{Descriptive statistics of the preprocessed corpus. Note: A tweet can have multiple sentences.}
    \label{corpus_stats}
\end{table}

\subsection{Unsupervised pre-training}
\label{unsup-pre-training}

\subsubsection{Architecture and pre-training procedure}
CrisisTransformers use the same architecture as $BERT_{BASE}$. In contrast to existing studies \cite{lee2020biobert,yang2020finbert,huang2019clinicalbert,nguyen2020bertweet}, we adopted a more versatile approach to selecting a pre-training procedure for our models. Instead of starting with a specific pre-training procedure, we experimented with multiple state-of-the-art models, namely MPNet, BERTweet, BERT, RoBERTa, XLM-RoBERTa, ALBERT, and ELECTRA, on classification tasks using 18 crisis-related labelled datasets (detailed in Section \ref{classification-datasets}). We observed RoBERTa and BERTweet emerging as the top-performing models on average. Therefore, we selected RoBERTa's pre-training procedure for training CrisisTransformers. Due to the extensive adoption of BERT and RoBERTa, we do not provide an in-depth explanation of the architecture in this paper; for more comprehensive insights, please refer to \cite{devlin-etal-2019-bert,liu2019roberta}. The configurations of the proposed models are provided in \ref{config}.

\subsubsection{Pre-training data}
We trained a Byte-Level BPE (Byte-Pair Encoding) tokenizer using the \textit{Tokenizers} library \cite{wolf2019huggingface} for our domain, utilizing the pre-processed crisis corpus discussed in Section \ref{crisis-corpus}. Acknowledging the nuanced nature of social media texts (the crisis corpus had 36 million unique tokens), we also set the vocabulary size to 64k \cite{nguyen2020bertweet}. Next, we used the trained tokenizer to tokenize the crisis corpus, thus generating sequence blocks of size $128$, on which we trained the CrisisTransformers. Table \ref{common_tokens} provides a comparative analysis of token counts in the vocabularies of established pre-trained models and CrisisTransformers. Among the existing models, RoBERTa and BERTweet share the highest similarity in vocabulary with CrisisTransformers.

\begin{table}
\scriptsize
    \centering
    \begin{tabular}{c|c|c}
    \hline
       \textbf{model}  & \textbf{intersection} & \textbf{unique}\\
    \hline \hline
       RoBERTa & 37,338 & 12,927\\
       BERTweet  & 15,121 & 48,880\\
       BERT & 7,905 & 21,091 \\
       XLM-RoBERTa & 6,431 & 243,571 \\
       MPNet & 5,754 & 24,773\\
       ELECTRA & 5,749 & 24,773\\
       ALBERT & 4,394 & 25,606\\
       \hline
    \end{tabular}
    \caption{Vocabulary similarity between existing pre-trained models and CrisisTransformers. Note: \textit{intersection} denotes the number of tokens shared between the existing models and CrisisTransformers, while \textit{unique} indicates the tokens exclusive to the vocabulary of the existing models.}
    \label{common_tokens}
\end{table}

\subsubsection{Optimization}

\begin{figure}
    \centering
    \includegraphics[width=0.9\textwidth]{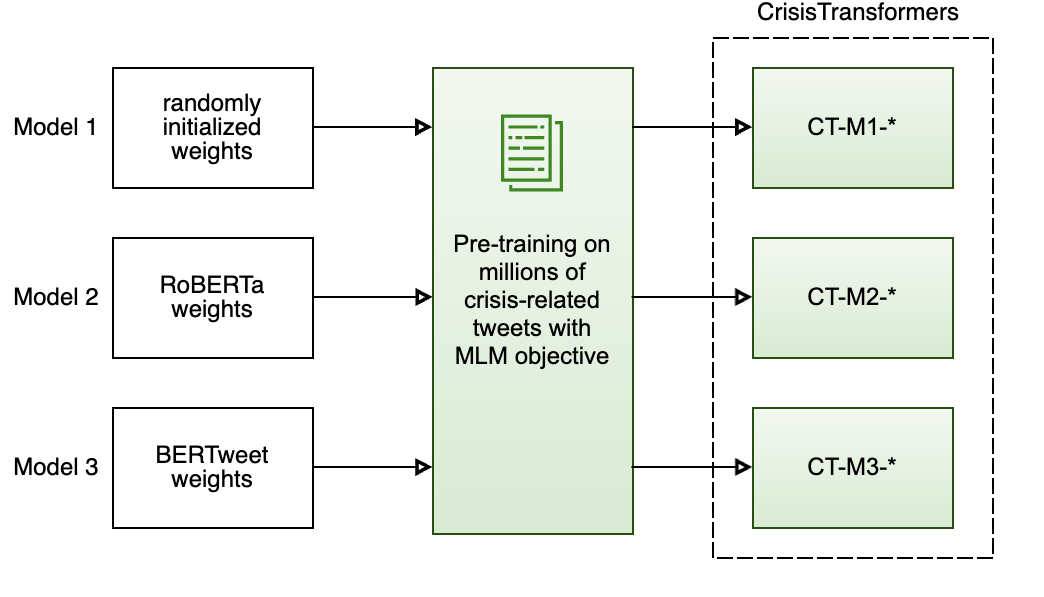}
    \caption{Pre-training of CrisisTransformers. Note: ``\texttt{*}" represents different checkpoints, which will be discussed later in Section \ref{results}.}
    \label{pre_training}
\end{figure}

We pre-trained three models (as shown in Figure \ref{pre_training}), utilizing 6 NVIDIA A100 GPUs (each with 80GB of memory). The training configurations for these models were as follows: CT-M1 (or \textbf{C}risis\textbf{T}ransformer-\textbf{M}odel\textbf{1}) was pre-trained from scratch with randomly initialized weights; CT-M2 had weights initialized with pre-trained RoBERTa's weights; and CT-M3 had weights initialized with pre-trained BERTweet's weights. CT-M1 was trained for 40 epochs, while CT-M2 and CT-M3 were trained for 20 epochs each. We used the \textit{Transformers} library \cite{wolf2019huggingface} to implement these models. 

For optimization, we employed the \textit{AdamW} optimizer with a peak learning rate set to 0.0004. To utilize the available GPU memory efficiently, we used a batch size of 8k with gradient accumulation steps of 16. Additionally, we set 5\% of the total training steps for warming up the learning rate. All three models finished training in two months.

\subsection{Fine-tuning}
\label{finetuning}

For fine-tuning the baselines and CrisisTransformers for text classification, as outlined in \cite{nguyen2020bertweet}, we added a linear prediction layer to the pooled output. We implemented mean pooling over the token embeddings of an input sequence while considering the attention mask. Both baselines and CrisisTransformers were fine-tuned under identical conditions. Each model was fine-tuned across 18 labelled crisis-related datasets for a maximum of 30 epochs, a batch size of 32, a learning rate of 1e-5, and AdamW as an optimizer. During each epoch, classification performance was assessed on a validation set. Early stopping was configured with a patience of 5 and a threshold of 0.0001. The final checkpoint was then used for evaluation on a test set. The fine-tuning procedure was repeated 5 times per model and dataset, with average performance scores being reported at a 95\% confidence interval.

\subsubsection{Labelled crisis-related datasets}
\label{classification-datasets}

Table \ref{labelled_datasets} lists the datasets we considered to evaluate both baselines and CrisisTransformers. Evaluating the performance across such diverse datasets from the crisis informatics domain was essential to test the robustness of our proposed models. CrisisBench \cite{alam2021crisisbench} provided the train/validation/test splits for datasets D-01 through D-06. For the remaining datasets, we implemented stratified sampling, allocating 70\% for training, 10\% for validation, and 20\% for testing, using scikit-learn's \textit{train-test split}\footnote{https://scikit-learn.org} with a random state of 42.

\begin{table}
    \centering
    \scriptsize
    \begin{tabular}{c|c|p{7cm}|c}
    \hline
      \textbf{Id}  & \textbf{Dataset} & \textbf{Description -- (\# of Classes)} & \textbf{Samples}\\
      \hline \hline
D-01 & CrisisMMD \cite{alam2018crisismmd} & Tweets from 7 disaster
events from 2017 -- (6) & 10,070\\
\hline
D-02 & CrisisLex \cite{olteanu2014crisislex} & Tweets from 26 different crisis events in 2012--13 -- (6) & 10,041 \\
\hline
D-03 & AIDR \cite{imran2014aidr} & Tweets collected by AIDR system -- (9) & 5,169\\
\hline
D-04 & ISCRAM2013 \cite{imran2013extracting} & Tweets from 2 different events in 2011 -- (5) & 810\\
\hline
D-05 & SWDM2013 \cite{imran2013practical} & Tweets related to Joplin tornado and Hurricane Sandy -- (4) & 346 \\
\hline
D-06 & CrisisNLP \cite{imran2016twitter} & Tweets from 19 different disaster events in 2013--15 -- (8) & 10,214\\
\hline
D-07 & Poddar et al. (2022) \cite{poddar2022winds} & Tweets related to stance towards COVID-19 vaccines -- (3) & 3,300 \\
\hline
D-08 & SAD Stressor \cite{mauriello2021sad} & SMS-like sentences mentioning everyday stressors & 6,850\\
\hline
D-09 & SAD Stress \cite{mauriello2021sad} & Stress and non-stress SMS-like sentences -- (2) & 6,850\\
\hline
D-10 & SAD COVID \cite{mauriello2021sad} & COVID and non-COVID SMS-like sentences -- (2) & 6,850\\
\hline
D-11 & LocBERT \cite{lamsal2022did} & COVID-19 tweets with origin and non-origin locations -- (2) & 2,800\\
\hline
D-12 & HMC (a) \cite{biddle2020leveraging} & Figurative versus literal health reports on Twitter -- (3) & 13,017\\
\hline
D-13 & Cotfas et al. (2021) \cite{cotfas2021longest} & Twitter opinions regarding COVID-19 vaccination -- (3) & 2,393\\
\hline
D-14 & HMC (b) \cite{biddle2020leveraging} & Disease mentions on tweets -- (10) & 13,017\\
\hline
D-15 & PHM \cite{karisani2018did} & Health mentions in social media -- (4) & 4,419\\
\hline
D-16 & Klein et al. (2021) (a) \cite{klein2021toward} & Tweets about actual and potential COVID-19 patients -- (3) & 4,266\\
\hline 
D-17 & Klein et al. (2021) (b) \cite{klein2021toward} & Tweets about groups of potential COVID-19 positive contacts -- (8) & 4,266\\
\hline
D-18 & ANTiVax \cite{hayawi2022anti} & Tweets on vaccine misinformation -- (2) & 11,518\\
\hline      
    \end{tabular}
    \caption{Labelled crisis datasets considered in this study for evaluating the performance of baselines and CrisisTransformers.}
    \label{labelled_datasets}
\end{table}

\subsection{Enriching sentence encoding}
\label{sentence-encoders}
By default, CrisisTransformers do not produce semantically rich embeddings, even though they were trained on a domain-specific corpus. Such pre-trained models require additional fine-tuning to learn to represent semantically similar sentences closer together within the vector space. These enhanced embeddings, capable of capturing semantic meanings, can then be effectively compared using cosine similarity. Their significance becomes particularly evident in tasks involving semantic search and clustering.

Our sentence encoders (CT-M1-*-SE, CT-M2-*-SE, and CT-M3-*-SE~---~where, ``SE" stands for \textbf{S}entence \textbf{E}ncoder) are built upon the recent success of utilizing siamese and triplet networks on sentence pairs \cite{conneau2017supervised} with pre-trained transformers \cite{reimers2019sentence} while leveraging the idea that adding corresponding contradicting pairs as ``hard negatives" alongside in-batch negatives further improves the performance \cite{gao2021simcse}. Expanding upon the method introduced in \cite{gao2021simcse}, we adapt it to utilize domain-specific pre-trained models instead of the existing general pre-trained models like BERT and RoBERTa. We used the following contrastive learning objectives to train our sentence encoders:

\begin{itemize}
    \item \textbf{Multiple Negative Ranking (MNR)}: This loss incorporates the (anchor, positive) pairs. Given a batch of pairs $(a_1, a^{+}_1), (a_2, a^{+}_2), \ldots, (a_n, a^{+}_n)$ where $(a_i, a^{+}_i)$ are positive pairs and $(a_i, a^{+}_j)$ for $i \neq j$ are considered negative pairs. The training objective for $(a_i, a^{+}_i)$ with mini-batch $N$ is:

\begin{equation}
l_i = - \log\left(\frac{e^{\text{similarity}(r_i, r^{+}_i)/\tau}}{\sum_{j=1}^{N} e^{\text{similarity}(r_i, r^{+}_j)/\tau}}\right)
\label{MNR}
\end{equation}

where, $r_i$ and $r^{+}_i$ are embeddings of $a_i$ and $a^{+}_i$ generated by our CrisisTransformers, \textit{similarity$(r_i, r^{+}_i)$} is cosine similarity, and $\tau$ is temperature hyperparameter.
    
    \item \textbf{MNR with hard negatives}: This loss incorporates the (anchor, positive, hard negative) pairs, i.e., $(a_n, a^{+}_n, a^{-}_n)$. The training objective in Equation \ref{MNR} can be modified to:

\begin{equation}
l_i = - \log\left(\frac{e^{\text{similarity}(r_i, r^{+}_i)/\tau}}{\sum_{j=1}^{N} (e^{\text{similarity}(r_i, r^{+}_j)/\tau} + e^{\text{similarity}(r_i, r^{-}_j)/\tau})}\right)
\label{mnr-negative}
\end{equation}
\end{itemize}

\begin{figure}
    \centering
    \includegraphics[width=0.9\textwidth]{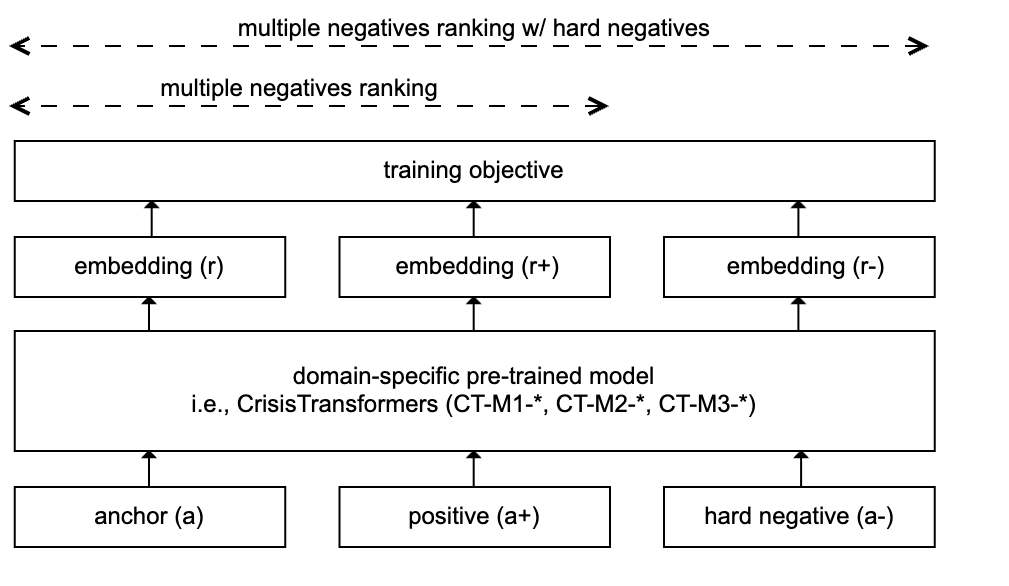}
    \caption{Training of our sentence encoders.}
    \label{se_training}
\end{figure}

The MNR loss maximizes the similarity between an anchor sentence and its positive sentence while considering all other positives in a batch as negatives. In MNR with hard negatives, the similarity between an anchor sentence and its positive sentence is maximized while using its hard negative and all other positive sentences in the same batch as negatives. We include the MNR training objective in the experiments for comparison purposes, even though MNR with hard negatives has been shown to outperform it \cite{gao2021simcse}. We train our sentence encoders (as shown in Figure \ref{se_training}) with these two objectives on (Question, Answer) pairs from GooAQ \cite{khashabi2021gooaq}, (anchor, positive, hard negative) triplets from QQP\footnote{https://quoradata.quora.com/First-Quora-Dataset-Release-Question-Pairs} \cite{embeddings-data} and (anchor, entailment, contradiction) triplets from AllNLI \cite{embeddings-data,bowman2015large,N18-1101} with a large batch size of 512 for a maximum of 20 epochs. We utilize a learning rate of 2e-05 and allocate 1\% of the total training steps for warm-up.

We implemented mean pooling over the token embeddings with attention to generate sentence embeddings.

\subsection{Evaluation setup}
\subsubsection{Classification task}
In line with prior research \cite{nguyen2020bertweet,beltagy-etal-2019-scibert}, we evaluate baselines and CrisisTransformers for the classification task using F1-macro, which considers the precision and recall of each class and provides an overall evaluation of the models' classification performance. For each dataset, we compute the F1-macro score as follows:

\begin{align*}
P_{\text{class}_i} &= \frac{TP_{\text{class}_i}}{TP_{\text{class}_i} + FP_{\text{class}_i}} \\
R_{\text{class}_i} &= \frac{TP_{\text{class}_i}}{TP_{\text{class}_i} + FN_{\text{class}_i}} \\
F1_{\text{class}_i} &= \frac{2 \cdot P_{\text{class}_i} \cdot R_{\text{class}_i}}{P_{\text{class}_i} + R_{\text{class}_i}} \\
{\text{F1-macro}} &= \frac{1}{N_{\text{classes}}} \sum_{i=1}^{N_{\text{classes}}} F1_{\text{class}_i}
\end{align*}

where, $TP_{\text{class}_i}$ is the number of true positive predictions for class $i$, $FP_{\text{class}_i}$ is the number of false positive predictions for class $i$, $FN_{\text{class}_i}$ is the number of false negative predictions for class $i$, and $N_{classes}$ is the total number of classes in the dataset.

\subsubsection{Sentence encoding task}

There is an absence of standard benchmark datasets in the crisis informatics domain to assess the semantic quality of the generated embeddings. In agreement with \cite{reimers2019sentence} and \cite{gao2021simcse} that the primary objective of the embeddings is to capture and represent semantic relationships in text data, we designed an alternative task. Our evaluation method involved calculating the weighted average cosine similarity among encoded tweets within individual classes in a labelled dataset, thereby measuring the semantic consistency of tweets belonging to the same class. This approach allowed us to capture the complexities and semantics of crisis-related content, resulting in a more insightful evaluation of the sentence embeddings.

\textbf{Task definition}:
Let $N$ represent the total number of crisis-related tweets in a dataset and $K$ denote the number of unique classes within the dataset. Let $E$ be a matrix of sentence embeddings, where each row $\mathbf{e}_i$ corresponds to the normalized embedding of the $i$-th tweet. Additionally, let $y$ be a vector containing the class labels associated with each tweet.

For each unique class $c_k$, the class weight $w_k$ is computed as the inverse of the count of tweets belonging to that class:

\[
w_k = \frac{1}{\text{count}(c_k)}
\]

These class weights are then normalized to obtain $\hat{w}_k$:

\[
\hat{w}_k = \frac{w_k}{\sum_{i=1}^{K} w_i}
\]

For each unique class $c_k$, the intra-class cosine similarity $d_k$ is computed. For each tweet $\mathbf{e}_i$ within class $c_k$, the average cosine similarity to other tweets within the same class is determined:

\[
d_k = \frac{1}{|\{i : y_i = c_k\}|} \sum_{i : y_i = c_k} \text{similarity}(\mathbf{e}_i, \mathbf{e}_j)
\]

Here, $\text{similarity}(\mathbf{e}_i, \mathbf{e}_j)$ calculates the cosine similarity between tweet embeddings $\mathbf{e}_i$ and $\mathbf{e}_j$, where $\mathbf{e}_j$ is a tweet within the same class as $\mathbf{e}_i$.

The weighted average distance $D_{\text{avg}}$ is computed across all classes, considering their respective normalized class weights $\hat{w}_k$:

\[
D_{\text{avg}} = \sum_{k=1}^{K} \hat{w}_k \cdot d_k
\]

$D_{\text{avg}}$ quantifies the average within-class semantic similarity of crisis-related tweets while accounting for the distribution of class weights.

The cosine similarity between sentence embeddings reflects how semantically similar or related the sentences are. If the embeddings are better at capturing the semantic content of crisis-related tweets within each class, the cosine similarity values within a class would be high. A higher cosine similarity within each class indicates that the embeddings effectively represent tweets that share similar content or context related to a specific crisis-related class. In summary, the higher the value of $D_{avg}$, the better the performance of a sentence encoder. We considered all the datasets listed in Table \ref{labelled_datasets} for this task.

\section{Results and Discussion}
\label{results}

\subsection{Checkpoints and convergence}
After the pre-training, we were interested in multiple checkpoints of CrisisTransformers: CT-M1-*, CT-M2-*, and CT-M3-*. CT-M1 was built from scratch and had two variants, CT-M1-BestLoss, representing the model at its lowest loss during training, and CT-M1-Complete, representing the model after 40 epochs. On the other hand, CT-M2 and CT-M3 were initialized using weights from pre-trained RoBERTa and BERTweet, respectively, and were trained up to 20 epochs each. CT-M2-OneLook represents the model after 1 epoch, while CT-M2-BestLoss and CT-M2-Complete represent the model at its lowest loss and the model after 20 epochs, respectively. The same setup was applied to CT-M3 models. In total, CrisisTransformers has 8 variants based on different checkpoints of CT-M1, CT-M2, and CT-M3 models.

\begin{figure}
    \centering
    \includegraphics[width=1\textwidth]{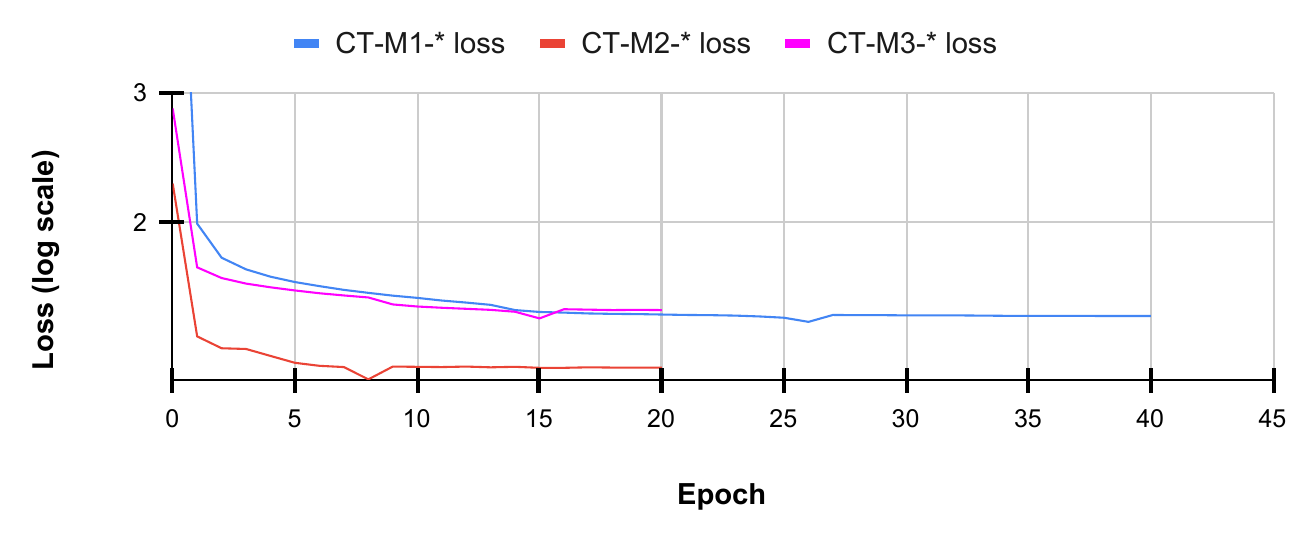}
    \caption{Validation loss versus epoch for CrisisTransformers' CT-M1-*, CT-M2-*, and CT-M3-* checkpoints, showing the impact of different initializations. The loss for CT-M1 at Epoch 0 was $9.841$, and it achieved its lowest loss at the 26th epoch. For CT-M2, the loss at Epoch 0 was 2.26, and it achieved its lowest loss at the 8th epoch. Lastly, CT-M3 started with a loss of 2.856 at Epoch 0 and reached its lowest loss at the 15th epoch. The y-axis is truncated to a maximum value of 3 for clarity. Although the data extends to $9.841$ on the y-axis, focusing on the range up to 3 enhances the visibility of differences between the plots, which may otherwise be overshadowed by the scale.}
    \label{loss}
\end{figure}

Figure \ref{loss} visualizes the validation loss versus epoch for each model. The graph provides insights into the impact of different initialization on the models' convergence. The loss patterns of the three models revealed distinct behaviours. CT-M1 demonstrated a gradual and consistent reduction in loss throughout the training period, suggesting steady convergence. CT-M2, on the other hand, exhibited a sharp initial drop in the loss within a few training steps, indicating rapid convergence and a smoother decline. Similarly, CT-M3 also displayed a significant initial loss drop. While CT-M3 initially shared a sharp loss drop with CT-M2, its convergence pattern aligned more with CT-M1 in the later epochs. The final loss of CT-M3 ultimately converged closer to that of CT-M1. All models seemed to plateau in their loss during the later epochs, indicating a potential convergence point. These loss patterns highlight the influence of different initializations on the time and trajectory of loss convergence; the pre-trained models seem to leverage their existing knowledge for a more efficient initial convergence than the model whose weights were randomly initialized.

\subsection{Evaluations}

For the classification task, we considered MPNet, BERTweet, BERT, RoBERTa, XLM-RoBERTa, ALBERT, and Electra as baselines for CrisisTransformers. As discussed in Section \ref{finetuning}, we finetuned the baselines and CrisisTransformers for the classification task across 18 different crisis-related datasets, each identified by a unique identifier (D-01 through D-18) (refer to Table \ref{labelled_datasets}). Results from the experiments are summarized in Table \ref{classification_results}.

Amongst the baselines, RoBERTa consistently outperforms other models with high F1-macro scores across several datasets. However, with the introduction of CrisisTransformers, the checkpoints of CT-M1 and CT-M2 stand out; overall, CrisisTransformers outperform the existing pre-trained models across all 18 datasets. The following models outperformed others in the respective number of datasets: CT-M1-BestLoss (in 4 datasets), CT-M1-Complete (in 3 datasets), CT-M2-OneLook (in 4 datasets), CT-M2-BestLoss (in 1 dataset), CT-M2-Complete (in 4 datasets), and CT-M3-OneLook (in 2 datasets). These results confirm the potential of CrisisTransformers for generalization and applicability in various crisis text classification tasks, which is particularly valuable for real-world applications such as disaster response, emergency communication, and crisis management.

Next, we performed sentence encoding tasks across all 18 datasets with the existing pre-trained models, CrisisTransformers, Sentence-Transformers, SimCSE, and CrisisTransformers-based sentence encoders. The results from the sentence encoding task are summarized in Tables \ref{cosine_pretrained}--\ref{cosine_se_2}.

The pre-trained models do not yield semantically meaningful sentence embeddings out-of-the-box. Nevertheless, one of our objectives was to investigate how well domain-adapted models perform in generating semantically meaningful sentence embeddings. Results show that (refer to Table \ref{cosine_pretrained}), within the existing pre-trained models, BERTweet emerged as a robust performer, consistently achieving competitive weighted average cosine similarity scores. However, CrisisTransformers, particularly the CT-M3 variants, invariably achieved the highest or second-highest scores regardless of the specific configuration (OneLook, BestLoss, or Complete). The performance of BERTweet and CT-M3 variants can be attributed to BERTweet's pre-training on an extensive corpus of tweets.  The results further indicate that the performance of the pre-trained RoBERTa is subpar. Consequently, the CT-M2 variants do not notably enhance performance. In contrast, the CT-M1 variants demonstrate a significant performance advantage over the CT-M2 variants. These findings suggest that further pre-training a domain-specific model on a sub-domain corpus (where ``tweets" reflect ``domain" and ``crisis-related tweets" indicate ``sub-domain") leads to improved performance in generating better sentence embeddings.

Furthermore, we trained CrisisTransformers using siamese and triplet networks with MNR and MNR with hard negatives training objectives, as discussed in Section \ref{sentence-encoders}, to create sentence encoders specifically designed for crisis-related social media texts. We used GooAQ (Question, Answer) pairs for MNR and QQP (anchor, positive, hard negative) triplets for MNR with hard negatives. As baselines for our sentence encoders, we utilized Sentence-Transformers and SimCSE. We considered the ``all-mpnet-base-v2" model\footnote{https://huggingface.co/sentence-transformers/all-mpnet-base-v2}, which is the highest-performing model in Sentence-Transformers, and the ``sup-simcse-roberta-base" model\footnote{https://huggingface.co/princeton-nlp/sup-simcse-roberta-base}, a high-performing base architecture model for SimCSE. We used only the first 10k pairs from GooAQ and QQP for both training objectives. We explored different sample sizes and ultimately found that using 10k pairs balanced model performance and having fewer training samples. This contrasts our baselines, where Sentence-Transformers was trained on over 1 billion samples, and SimCSE was trained on 314k samples. 

Table \ref{cosine_se_1} and Table \ref{cosine_se_2} summarize the performance of the baselines and our sentence encoders in terms of the weighted average cosine similarity, and Table \ref{average_results} reports the overall performance. Across all 18 datasets, our sentence encoders outperform both Sentence-Transformers and SimCSE. Notably, CT-M1-Complete-SE (MNR) and CT-M2-Complete-SE (hard negatives) each achieved the best performances across 4 datasets, and CT-M1-BestLoss-SE (MNR) and CT-M2-BestLoss-SE (hard negatives) each in 3 datasets. Overall, CT-M1 variants performed better in 11 datasets, CT-M2 variants in 6 datasets, and CT-M3 in 1 dataset. Considering training objectives, models trained with hard negatives achieved the highest scores across 11 datasets. CT-M1-Complete-SE (hard negatives), although trained on 10k samples, achieved an average score of 0.7140, surpassing the current state-of-the-art by 12\% while outperforming Sentence-Transformers' average score of 0.6374. These results highlight the adaptability and effectiveness of CrisisTransformers-based sentence encoders in capturing semantic similarity within sentences, particularly in crisis-related contexts. This reinforces the idea that tailoring models to specific domains, like crisis situations, can yield significant improvements over more general-purpose models in sentence encoding tasks, even when trained with less data. Among the baselines, Sentence-Transformers performed better compared to SimCSE across all datasets. In fact, our CT-M3 variants (avg. scores ranging from 0.2663 to 0.2792) outperformed SimCSE (avg. score of 0.1765). The noticeable performance advantage of Sentence-Transformers over SimCSE can be attributed to the comprehensive training of its ``all-mpnet-base-v2'' model, which involved training on more than 1 billion sentence pairs/triplets. This extensive training likely provided the model with a broader and richer understanding of general language nuances, thus contributing to its superior performance.

Motivated to study the effect of training samples, we re-trained CT-M1-Complete-SE (hard negatives) while increasing the training samples from 10k to 102k samples (complete QQP) and further augmented the AllNLI dataset to create a training size of 378k. After this re-training, we observed an improvement of approx. 3.56\% with complete QQP and approx. 4.83\% with QQP+AllNLI. Overall, our best-performing sentence encoder improved the current state-of-the-art by around 17.43\%. This observation sets the stage for potential enhancements to our sentence encoder. Going forward, our future objectives include training our sentence encoders on a scale similar to Sentence-Transformers for an even more substantial improvement.

\begin{landscape}
\begin{table}
\scriptsize
\centering

\hspace*{-3cm}\begin{tabular}{p{2.3cm}|p{2.1cm}p{2.1cm}p{2.1cm}p{2.1cm}p{2.1cm}p{2.1cm}p{2.1cm}p{2.1cm}p{2.1cm}}
\hline
\textbf{model} & \textbf{D-01}          & \textbf{D-02}          & \textbf{D-03}          & \textbf{D-04}          & \textbf{D-05}         & \textbf{D-06}          & \textbf{D-07}          & \textbf{D-08}          & \textbf{D-09}          \\
\hline \hline
MPNet          & 0.6559 ±0.005          & 0.7571 ±0.004          & 0.4286 ±0.036          & 0.7359 ±0.000          & 0.5706 ±0.168         & 0.7776 ±0.002          & 0.5810 ±0.014           & 0.6934 ±0.022          & 0.6758 ±0.000          \\
BERTweet       & 0.6597 ±0.024          & 0.7569 ±0.001          & 0.5735 ±0.022          & 0.6979 ±0.024          & 0.6574 ±0.183         & 0.7871 ±0.002          & 0.5890 ±0.010           & 0.7148 ±0.002          & 0.6940 ±0.004           \\
BERT           & 0.6728 ±0.026          & 0.7297 ±0.000          & 0.5956 ±0.008          & 0.6089 ±0.084          & 0.6495 ±0.078         & 0.7782 ±0.001          & 0.5197 ±0.013          & 0.7040 ±0.002           & 0.6770 ±0.000           \\
RoBERTa        & 0.6891 ±0.023          & 0.7603 ±0.002          & 0.6240 ±0.000           & 0.7650 ±0.021           & 0.7683 ±0.014         & 0.7919 ±0.000          & 0.5808 ±0.010          & 0.7171 ±0.002          & 0.7122 ±0.008          \\
XLM-RoBERTa    & 0.6419 ±0.006          & 0.7505 ±0.003          & 0.5908 ±0.020          & 0.6321 ±0.049          & 0.3849 ±0.127         & 0.7918 ±0.001          & 0.5187 ±0.015          & 0.7232 ±0.001          & 0.6702 ±0.000          \\
ALBERT         & 0.6548 ±0.038          & 0.7281 ±0.006          & 0.5571 ±0.022          & 0.5589 ±0.070          & 0.4484 ±0.023         & 0.7648 ±0.005          & 0.4963 ±0.020          & 0.6982 ±0.016          & 0.6989 ±0.000          \\
ELECTRA        & 0.6427 ±0.008          & 0.7409 ±0.000          & 0.5766 ±0.049          & 0.5422 ±0.036          & 0.2685 ±0.045         & 0.7874 ±0.003          & 0.5699 ±0.009          & 0.7183 ±0.005          & 0.7191 ±0.010          \\
\hline
CT-M1-BestLoss & 0.6555 ±0.002          & 0.7539 ±0.002          & 0.6174 ±0.041          & 0.6730 ±0.033           & \textbf{0.8510 ±0.018} & \textbf{0.7927 ±0.004} & \textbf{0.7118 ±0.009} & 0.7121 ±0.003          & 0.7023 ±0.012          \\
CT-M1-Complete & 0.6567 ±0.000          & 0.7613 ±0.000          & 0.6291 ±0.003          & 0.6870 ±0.004           & 0.8160 ±0.018          & 0.7703 ±0.032          & 0.7035 ±0.009          & 0.709 ±0.008           & 0.6980 ±0.005           \\
CT-M2-OneLook  & \textbf{0.6916 ±0.032} & 0.757 ±0.000           & \textbf{0.6651 ±0.009} & \textbf{0.7744 ±0.040} & 0.8274 ±0.024         & 0.7862 ±0.000          & 0.6504 ±0.005          & 0.7250 ±0.000           & 0.7046 ±0.003          \\
CT-M2-BestLoss & 0.6646 ±0.013          & 0.7643 ±0.001          & 0.6637 ±0.007          & 0.7669 ±0.014          & 0.7874 ±0.074         & 0.7860 ±0.006           & 0.6721 ±0.003          & 0.7207 ±0.000          & 0.6969 ±0.000          \\
CT-M2-Complete & 0.6606 ±0.008          & \textbf{0.7656 ±0.000} & 0.6569 ±0.013          & 0.7453 ±0.019          & 0.8317 ±0.006         & 0.7796 ±0.005          & 0.6621 ±0.012          & 0.7015 ±0.014          & 0.6889 ±0.013          \\
CT-M3-OneLook  & 0.6494 ±0.002          & 0.7592 ±0.000          & 0.5383 ±0.052          & 0.7021 ±0.015          & 0.8048 ±0.034         & 0.7779 ±0.005          & 0.6587 ±0.010          & \textbf{0.7291 ±0.002} & \textbf{0.7445 ±0.005} \\
CT-M3-BestLoss & 0.6546 ±0.006          & 0.7585 ±0.000          & 0.5555 ±0.054          & 0.7165 ±0.030          & 0.5725 ±0.285         & 0.7874 ±0.000          & 0.6729 ±0.003          & 0.7173 ±0.004          & 0.7200 ±0.000            \\
CT-M3-Complete & 0.6547 ±0.004          & 0.7590 ±0.002           & 0.5726 ±0.053          & 0.7011 ±0.011          & 0.6806 ±0.189         & 0.7898 ±0.002          & 0.6814 ±0.004          & 0.7156 ±0.005          & 0.7061 ±0.032\\
\hline
\end{tabular}

\vspace{1cm}
\hspace*{-3cm}\begin{tabular}{p{2.3cm}|p{2.1cm}p{2.1cm}p{2.1cm}p{2.1cm}p{2.1cm}p{2.1cm}p{2.1cm}p{2.1cm}p{2.1cm}}
\hline
\textbf{model}          & \textbf{D-10}   & \textbf{D-11}   & \textbf{D-12}   & \textbf{D-13}   & \textbf{D-14}   & \textbf{D-15}   & \textbf{D-16}   & \textbf{D-17}  & \textbf{D-18}   \\
\hline \hline
MPNet          & 0.9208 ±0.002          & 0.7590 ±0.000           & 0.8882 ±0.003          & 0.8119 ±0.006          & 0.9905 ±0.000         & 0.8029 ±0.000          & 0.8040 ±0.009           & 0.5260 ±0.008       & 0.9829 ±0.000          \\
BERTweet       & 0.9358 ±0.004          & 0.7727 ±0.005          & 0.9009 ±0.002          & 0.8562 ±0.000          & 0.9933 ±0.000         & 0.8209 ±0.010          & 0.8108 ±0.016          & 0.5274 ±0.001      & 0.9830 ±0.001           \\
BERT           & 0.9001 ±0.009          & 0.7230 ±0.008           & 0.8745 ±0.000          & 0.7595 ±0.000          & 0.9899 ±0.000         & 0.8106 ±0.003          & 0.7662 ±0.010          & 0.5941 ±0.018      & 0.9748 ±0.002          \\
RoBERTa        & 0.9125 ±0.002          & 0.7665 ±0.005          & 0.8904 ±0.000          & 0.8338 ±0.005          & 0.9927 ±0.000         & 0.8310 ±0.006           & 0.7998 ±0.005          & 0.6182 ±0.050       & 0.9837 ±0.000          \\
XLM-RoBERTa    & 0.9167 ±0.004          & 0.7703 ±0.003          & 0.8835 ±0.003          & 0.7975 ±0.000          & 0.9930 ±0.000          & 0.8101 ±0.004          & 0.7902 ±0.008          & 0.5280 ±0.041       & 0.9773 ±0.003          \\
ALBERT         & 0.8759 ±0.000          & 0.7623 ±0.006          & 0.8750 ±0.001           & 0.7625 ±0.018          & 0.9917 ±0.000         & 0.8050 ±0.016           & 0.8011 ±0.011          & 0.5709 ±0.029      & 0.9760 ±0.002           \\
ELECTRA        & 0.9005 ±0.003          & 0.7820 ±0.007           & 0.8878 ±0.001          & 0.8268 ±0.004          & 0.9914 ±0.000         & 0.8119 ±0.009          & 0.8008 ±0.006          & 0.5027 ±0.016      & 0.9810 ±0.000           \\
\hline
CT-M1-BestLoss & 0.9336 ±0.007          & 0.7855 ±0.006          & 0.8988 ±0.002          & \textbf{0.8654 ±0.011} & 0.9929 ±0.000         & 0.8260 ±0.011           & 0.8537 ±0.007          & 0.5662 ±0.051      & 0.9885 ±0.000          \\
CT-M1-Complete & 0.9400 ±0.007            & 0.7817 ±0.008          & \textbf{0.9043 ±0.000} & 0.8641 ±0.006          & 0.9916 ±0.000         & 0.8343 ±0.004          & \textbf{0.8551 ±0.004} & 0.5573 ±0.060      & \textbf{0.9893 ±0.000} \\
CT-M2-OneLook  & 0.9337 ±0.003          & 0.7775 ±0.006          & 0.8988 ±0.003          & 0.8613 ±0.003          & 0.9928 ±0.000         & \textbf{0.8427 ±0.005} & 0.8407 ±0.008          & 0.7080 ±0.000           & 0.9851 ±0.000          \\
CT-M2-BestLoss & 0.9392 ±0.004          & 0.7875 ±0.002          & 0.8952 ±0.004          & 0.8490 ±0.005           & \textbf{0.9940 ±0.000} & 0.8265 ±0.003          & 0.8244 ±0.002          & 0.6304 ±0.044      & 0.9861 ±0.001          \\
CT-M2-Complete & \textbf{0.9476 ±0.000} & \textbf{0.7903 ±0.014} & 0.8978 ±0.000          & 0.8491 ±0.007          & 0.9930 ±0.000          & 0.8400 ±0.007            & 0.8207 ±0.002          & \textbf{0.7254 ±0.000} & 0.9881 ±0.001          \\
CT-M3-OneLook  & 0.9386 ±0.000          & 0.7861 ±0.003          & 0.8946 ±0.001          & 0.8268 ±0.014          & 0.9938 ±0.000         & 0.8235 ±0.005          & 0.8319 ±0.004          & 0.6026 ±0.006      & 0.9865 ±0.000          \\
CT-M3-BestLoss & 0.9439 ±0.000          & 0.7829 ±0.003          & 0.8968 ±0.000          & 0.8594 ±0.008          & 0.9938 ±0.000         & 0.8388 ±0.003          & 0.8328 ±0.005          & 0.6235 ±0.064      & 0.9864 ±0.000          \\
CT-M3-Complete & 0.9398 ±0.003          & 0.7785 ±0.005          & 0.8961 ±0.000          & 0.8510 ±0.002           & 0.9933 ±0.000         & 0.8103 ±0.003          & 0.8338 ±0.000          & 0.6470 ±0.046       & 0.9894 ±0.000         \\
\hline
\end{tabular}

\caption{Performance of the existing pre-trained models and CrisisTransformers (CT-*) on classification task across 18 crisis datasets (D-01 through D-18), with average F1-macro (at 95\% confidence interval) being reported. For the corresponding dataset names of each dataset identifier, please refer to Table \ref{labelled_datasets}. The best scores are shown in \textbf{bold}.}
\label{classification_results}
\end{table}
\end{landscape}

\begin{landscape}
\begin{table}
\centering
\scriptsize
\begin{tabular}{p{2.5cm}|p{1cm}p{1cm}p{1cm}p{1cm}p{1cm}p{1cm}p{1cm}p{1cm}p{1cm}}
\hline
\textbf{model}                 & \textbf{D-1}    & \textbf{D-2}    & \textbf{D-3}    & \textbf{D-4}    & \textbf{D-5}    & \textbf{D-6}    & \textbf{D-7}    & \textbf{D-8}    & \textbf{D-9}    \\
\hline \hline
MPNet                 & 0.0709 & 0.0777 & 0.0783 & 0.0795 & 0.073  & 0.0739 & 0.0618 & 0.0938 & 0.1719 \\
BERTweet              & 0.2532 & 0.2508 & \underline{0.2687} & 0.2349 & 0.218  & 0.2382 & 0.2255 & 0.2465 & 0.299  \\
BERT                  & 0.0897 & 0.1182 & 0.1108 & 0.121  & 0.1079 & 0.1076 & 0.1171 & 0.1904 & 0.2429 \\
RoBERTa               & 0.0235 & 0.0273 & 0.0272 & 0.0272 & 0.0252 & 0.0254 & 0.0276 & 0.037  & 0.0502 \\
XLM-RoBERTa           & 0.0026 & 0.0027 & 0.0028 & 0.0026 & 0.0024 & 0.0025 & 0.0024 & 0.0041 & 0.0074 \\
ALBERT                & 0.1129 & 0.1249 & 0.1262 & 0.1286 & 0.1088 & 0.1183 & 0.1225 & 0.1854 & 0.272  \\
ELECTRA               & 0.0646 & 0.0713 & 0.0805 & 0.0713 & 0.063  & 0.0665 & 0.0554 & 0.0898 & 0.1702 \\
\hline
CT-M1-BestLoss        & 0.1129 & 0.1238 & 0.1216 & 0.1117 & 0.1094 & 0.1243 & 0.1303 & 0.1137 & 0.1407 \\
CT-M1-Complete        & 0.1078 & 0.1186 & 0.1164 & 0.1064 & 0.1052 & 0.1199 & 0.127  & 0.1111 & 0.1368 \\
CT-M2-OneLook         & 0.0309 & 0.036  & 0.034  & 0.0346 & 0.0328 & 0.032  & 0.0378 & 0.0456 & 0.0625 \\
CT-M2-BestLoss        & 0.0527 & 0.0581 & 0.0548 & 0.0561 & 0.0515 & 0.052  & 0.0676 & 0.0766 & 0.1019 \\
CT-M2-Complete        & 0.0541 & 0.0584 & 0.0552 & 0.0564 & 0.0521 & 0.0529 & 0.0713 & 0.0731 & 0.097  \\
CT-M3-OneLook         & 0.2712 & \textbf{0.2962} & 0.2636 & \textbf{0.2694} & \textbf{0.2578} & 0.2648 & \textbf{0.2921} & \textbf{0.273}  & \textbf{0.3179} \\
CT-M3-BestLoss        & \textbf{0.2758} & \underline{0.2935} & \textbf{0.2697} & \underline{0.2579} & \underline{0.2468} & \textbf{0.2673} & \underline{0.2867} & \underline{0.253}  & \underline{0.3008} \\
CT-M3-Complete        & \underline{0.2732} & 0.2853 & 0.2679 & 0.2492 & 0.2404 & \underline{0.2661} & 0.2858 & 0.2432 & 0.2874 \\
\hline

\end{tabular}

\vspace{2cm}
\begin{tabular}{p{2.5cm}|p{1cm}p{1cm}p{1cm}p{1cm}p{1cm}p{1cm}p{1cm}p{1cm}p{1cm}}
\hline
\textbf{model}                 & \textbf{D-1}    & \textbf{D-2}    & \textbf{D-3}    & \textbf{D-4}    & \textbf{D-5}    & \textbf{D-6}    & \textbf{D-7}    & \textbf{D-8}    & \textbf{D-9}    \\
\hline \hline
MPNet                 & 0.1119 & 0.053  & 0.0961 & 0.0629 & 0.0962 & 0.0813 & 0.0654 & 0.054  & 0.0665 \\
BERTweet              & 0.2475 & 0.1997 & 0.3069 & 0.2261 & 0.3039 & 0.2776 & 0.2283 & 0.1949 & 0.2452 \\
BERT                  & 0.1839 & 0.102  & 0.1684 & 0.1076 & 0.162  & 0.1587 & 0.1423 & 0.1154 & 0.1309 \\
RoBERTa               & 0.0354 & 0.0246 & 0.0372 & 0.0262 & 0.0378 & 0.0331 & 0.0292 & 0.0239 & 0.0302 \\
XLM-RoBERTa           & 0.0042 & 0.0021 & 0.004  & 0.0026 & 0.004  & 0.0032 & 0.0033 & 0.0023 & 0.003  \\
ALBERT                & 0.1933 & 0.1109 & 0.1666 & 0.1246 & 0.1671 & 0.1516 & 0.1291 & 0.1069 & 0.1366 \\
ELECTRA               & 0.1134 & 0.0439 & 0.0922 & 0.0596 & 0.0916 & 0.0796 & 0.0665 & 0.0416 & 0.0648 \\
\hline
CT-M1-BestLoss        & 0.1188 & 0.1168 & 0.1454 & 0.1279 & 0.1451 & 0.1324 & 0.1173 & 0.0986 & 0.136  \\
CT-M1-Complete        & 0.1161 & 0.1135 & 0.1409 & 0.1248 & 0.1403 & 0.1285 & 0.115  & 0.0977 & 0.1328 \\
CT-M2-OneLook         & 0.0445 & 0.0317 & 0.0505 & 0.0353 & 0.0509 & 0.0444 & 0.0372 & 0.0314 & 0.0407 \\
CT-M2-BestLoss        & 0.078  & 0.0503 & 0.0791 & 0.0631 & 0.0786 & 0.0697 & 0.0627 & 0.0587 & 0.0724 \\
CT-M2-Complete        & 0.0744 & 0.0528 & 0.0791 & 0.0651 & 0.0788 & 0.0691 & 0.0641 & 0.0601 & 0.0751 \\
CT-M3-OneLook         & \textbf{0.2831} & \textbf{0.2644} & \textbf{0.3284} & \textbf{0.2814} & \textbf{0.3194} & \textbf{0.3028} & \textbf{0.2418} & \textbf{0.2107} & \textbf{0.2886} \\
CT-M3-BestLoss        & \underline{0.2586} & \underline{0.2537} & \underline{0.3224} & 0.2782 & \underline{0.3162} & \underline{0.2938} & \underline{0.2315} & \underline{0.1996}& 0.2829 \\
CT-M3-Complete        & 0.25   & 0.2519 & 0.3122 & \underline{0.2795} & 0.3071 & 0.2844 & 0.2283 & 0.1965 & \underline{0.2856} \\
\hline
\end{tabular}

\caption{Performance of the existing pre-trained models and CrisisTransformers on sentence encoding task across 18 crisis datasets (D-01 through D-18), with weighted average cosine similarity being reported. The best scores are shown in \textbf{bold}, and the second-best scores are \underline{underlined}. Note that results reported in this table are intended for comparative purposes only; embeddings generated by pre-trained models out-of-the-box do not produce semantically meaningful sentence embeddings.}
\label{cosine_pretrained}
\end{table}
\end{landscape}

\begin{landscape}
\begin{table}
\centering
\scriptsize
\begin{tabular}{l|lllllllll}
\hline
\textbf{model}        & \textbf{D-1}    & \textbf{D-2}    & \textbf{D-3}    & \textbf{D-4}    & \textbf{D-5}    & \textbf{D-6}    & \textbf{D-7}    & \textbf{D-8}    & \textbf{D-9}    \\ \hline \hline
Sentence-Transformers & 0.6103          & 0.6809          & 0.5407          & 0.5632          & 0.487           & 0.5698          & 0.6019          & 0.7528          & 0.8749          \\
SimCSE                & 0.18            & 0.18            & 0.15            & 0.17            & 0.16            & 0.17            & 0.17            & 0.19            & 0.23            \\ \hline
CT-M1-BestLoss-SE$_{\text{MNR}}$ & 0.6201          & \textbf{0.7647} & 0.5936          & 0.59            & \textbf{0.6276} & 0.6344          & 0.8147          & 0.7117          & 0.8438          \\
CT-M1-Complete-SE$_{\text{MNR}}$ & 0.6278          & \underline{0.7628}    & 0.5964          & 0.5904          & \underline{0.6229}    & 0.6382          & \underline{0.8212}    & 0.7156          & 0.8469          \\
CT-M2-OneLook-SE$_{\text{MNR}}$  & 0.5535          & 0.7057          & 0.5418          & 0.5652          & 0.5297          & 0.5594          & 0.7277          & 0.7149          & 0.8667          \\
CT-M2-BestLoss-SE$_{\text{MNR}}$ & 0.587           & 0.735           & 0.535           & 0.5968          & 0.5542          & 0.5794          & 0.8034          & 0.7136          & 0.8705          \\
CT-M2-Complete-SE$_{\text{MNR}}$ & 0.5834          & 0.7366          & 0.5391          & 0.5955          & 0.5573          & 0.5834          & 0.8002          & 0.71            & 0.8721          \\
CT-M3-OneLook-SE$_{\text{MNR}}$  & 0.5672          & 0.7155          & 0.5484          & 0.5756          & 0.5459          & 0.5709          & 0.7593          & 0.701           & 0.8275          \\
CT-M3-BestLoss-SE$_{\text{MNR}}$ & 0.5971          & 0.7265          & 0.5528          & 0.5864          & 0.5378          & 0.5922          & 0.778           & 0.6924          & 0.8215          \\
CT-M3-Complete-SE$_{\text{MNR}}$ & 0.6002          & 0.7262          & 0.5556          & 0.582           & 0.5382          & 0.5951          & 0.785           & 0.6937          & 0.825           \\ \hline
CT-M1-BestLoss-SE$_{\text{MNR~w/~hard~negatives}}$  & \textbf{0.6465} & 0.7338          & \underline{0.6059}    & 0.615           & 0.6016          & \textbf{0.6649} & 0.7874          & 0.7657          & 0.871           \\
CT-M1-Complete-SE$_{\text{MNR~w/~hard~negatives}}$  & \underline{0.6432}    & 0.7385          & \textbf{0.6097} & 0.6151          & 0.6053          & \underline{0.6626}    & 0.7912          & 0.7685          & \underline{0.8735}          \\
CT-M2-OneLook-SE$_{\text{MNR~w/~hard~negatives}}$   & 0.5702          & 0.6753          & 0.5536          & 0.5924          & 0.5465          & 0.5709          & 0.7063          & 0.7554          & 0.8733          \\
CT-M2-BestLoss-SE$_{\text{MNR~w/~hard~negatives}}$  & 0.5855          & 0.7145          & 0.5436          & 0.6123          & 0.6163          & 0.5903          & 0.8158          & \textbf{0.7784} & \textbf{0.8833}    \\
CT-M2-Complete-SE$_{\text{MNR~w/~hard~negatives}}$  & 0.5972          & 0.7208          & 0.5466          & 0.6116          & 0.6118          & 0.6049          & \textbf{0.8261} & \underline{0.7769}    & \textbf{0.8883} \\
CT-M3-OneLook-SE$_{\text{MNR~w/~hard~negatives}}$   & 0.605           & 0.7022          & 0.5583          & 0.6147          & 0.5674          & 0.595           & 0.7504          & 0.7301          & 0.828           \\
CT-M3-BestLoss-SE$_{\text{MNR~w/~hard~negatives}}$  & 0.6307          & 0.715           & 0.5687          & \textbf{0.6199} & 0.5529          & 0.6151          & 0.7652          & 0.7334          & 0.8289          \\
CT-M3-Complete-SE$_{\text{MNR~w/~hard~negatives}}$  & 0.6314          & 0.7144          & 0.5684          & \underline{0.6171}    & 0.5518          & 0.6161          & 0.7682          & 0.7307          & 0.8274    \\      \hline
\end{tabular}
\caption{(Part 1/2) Performance of baselines and our sentence encoders in the sentence encoding task. The best scores are shown in \textbf{bold}, and the second-best scores are \underline{underlined}.}
\label{cosine_se_1}
\end{table}
\end{landscape}

\begin{landscape}
\begin{table}
\centering
\scriptsize
\begin{tabular}{l|lllllllll}
\hline
\textbf{model}        & \textbf{D-10}    & \textbf{D-11}    & \textbf{D-12}    & \textbf{D-13}    & \textbf{D-14}    & \textbf{D-15}    & \textbf{D-16}    & \textbf{D-17}    & \textbf{D-18}    \\ \hline \hline
Sentence-Transformers & 0.6674                            & 0.7297        & 0.8211                            & 0.4939                            & 0.7259                            & 0.7579                            & 0.532                             & 0.5192                            & 0.545                             \\
SimCSE                & 0.19                              & 0.18          & 0.2                               & 0.15                              & 0.19                              & 0.19                              & 0.1683                            & 0.15                              & 0.16                              \\ \hline
CT-M1-BestLoss-SE$_{\text{MNR}}$ & 0.6878                            & \underline{0.8553}                      & \underline{0.8426}                      & 0.7094                            & \textbf{0.7609}                   & 0.7624                            & \underline{0.6788}                      & \underline{0.5874}                      & 0.727                             \\
CT-M1-Complete-SE$_{\text{MNR}}$ & 0.6904                            & \textbf{0.8584}                   & \textbf{0.8439}                   & 0.7146                            & \underline{0.7599}                      & 0.7567                            & \textbf{0.6814}                   & \textbf{0.5918}                   & 0.728                             \\
CT-M2-OneLook-SE$_{\text{MNR}}$  & 0.685                             & 0.7972                            & 0.823                             & 0.6105                            & 0.7357                            & 0.7403                            & 0.6182                            & 0.5505                            & 0.6392                            \\
CT-M2-BestLoss-SE$_{\text{MNR}}$ & 0.7075                            & 0.838                             & 0.8375                            & 0.7005                            & 0.7551                            & 0.7424                            & 0.6047                            & 0.5555                            & 0.6978                            \\
CT-M2-Complete-SE$_{\text{MNR}}$ & 0.7035                            & 0.8425                            & 0.8383                            & 0.6952                            & 0.7597                            & 0.7432                            & 0.6124                            & 0.5727                            & 0.6824                            \\
CT-M3-OneLook-SE$_{\text{MNR}}$  & 0.6633                            & 0.8135                            & 0.8097                            & 0.6797                            & 0.7247                            & 0.7391                            & 0.6155                            & 0.536                             & 0.6602                            \\
CT-M3-BestLoss-SE$_{\text{MNR}}$ & 0.6659                            & 0.8137                            & 0.8107                            & 0.6831                            & 0.7346                            & 0.7421                            & 0.6508                            & 0.5655                            & 0.6778                            \\
CT-M3-Complete-SE$_{\text{MNR}}$ & 0.6644                            & 0.8175                            & 0.8134                            & 0.6907                            & 0.7347                            & 0.7464                            & 0.652                             & 0.5657                            & 0.6857                            \\ \hline
CT-M1-BestLoss-SE$_{\text{MNR~w/~hard~negatives}}$  & \underline{0.7424}                      & 0.7961                            & 0.8232                            & 0.7455                            & 0.744                             & \underline{0.7867}                      & 0.6526                            & 0.5346                            & 0.727                             \\
CT-M1-Complete-SE$_{\text{MNR~w/~hard~negatives}}$  & 0.7417                            & 0.7912                            & 0.8205                            & 0.7496                            & 0.7441                            & \textbf{0.7868}                   & 0.6547                            & 0.531                             & 0.7261                            \\
CT-M2-OneLook-SE$_{\text{MNR~w/~hard~negatives}}$   & 0.7205                            & 0.7013                            & 0.7833                            & 0.626                             & 0.7207                            & 0.7593                            & 0.5885                            & 0.5022                            & 0.6529                            \\
CT-M2-BestLoss-SE$_{\text{MNR~w/~hard~negatives}}$  & \textbf{0.7478}                   & 0.7925                            & 0.8013                            & \underline{0.7532}                      & 0.7453                            & 0.7759                            & 0.6333                            & 0.5363                            & \underline{0.7323}                      \\
CT-M2-Complete-SE$_{\text{MNR~w/~hard~negatives}}$  & 0.7399                            & 0.7851                            & 0.8094                            & \textbf{0.7632}                   & 0.7564                            & 0.7815                            & 0.6416                            & 0.5417                            & \textbf{0.743}                    \\
CT-M3-OneLook-SE$_{\text{MNR~w/~hard~negatives}}$   & 0.704                             & 0.7404                            & 0.7825                            & 0.695                             & 0.7138                            & 0.7482                            & 0.6241                            & 0.524                             & 0.6728                            \\
CT-M3-BestLoss-SE$_{\text{MNR~w/~hard~negatives}}$  & 0.7092                            & 0.7419                            & 0.7885                            & 0.7032                            & 0.7235                            & 0.7641                            & 0.6355                            & 0.5378                            & 0.6883                            \\
CT-M3-Complete-SE$_{\text{MNR~w/~hard~negatives}}$  & 0.7071                            & 0.7442                            & 0.7894                            & 0.7049                            & 0.7241                            & 0.7635                            & 0.6361                            & 0.543                             & 0.6903                            \\                    \hline             
\end{tabular}
\caption{(Part 2/2) Performance of baselines and our sentence encoders in the sentence encoding task. The best scores are shown in \textbf{bold}, and the second-best scores are \underline{underlined}.}
\label{cosine_se_2}
\end{table}
\end{landscape}

\begin{table}[ht]
    \centering
    \scriptsize
    \begin{tabular}{l|l}
    \hline
    \textbf{sentence encoder} & \textbf{avg. score}\\
    \hline
    \hline
      Sentence-Transformers   &  0.6374\\
      SimCSE & 0.1765\\
      \hline
        \textit{10k training samples of GooAQ} &\\
    -- CT-M1-BestLoss-SE$_{\text{MNR}}$ & 0.7117\\
    -- CT-M1-Complete-SE$_{\text{MNR}}$ & 0.7137\\
    -- CT-M2-OneLook-SE$_{\text{MNR}}$ & 0.6646\\
    -- CT-M2-BestLoss-SE$_{\text{MNR}}$ & 0.6896\\
    -- CT-M2-Complete-SE$_{\text{MNR}}$ & 0.6904\\
    -- CT-M3-OneLook-SE$_{\text{MNR}}$ & 0.6696\\
    -- CT-M3-BestLoss-SE$_{\text{MNR}}$ & 0.6793\\
    -- CT-M3-Complete-SE$_{\text{MNR}}$ & 0.6817\\
      \hline
        \textit{10k training samples of QQP} &\\
    -- CT-M1-BestLoss-SE$_{\text{MNR~w/~hard~negatives}}$ & 0.7135\\
    -- CT-M1-Complete-SE$_{\text{MNR~w/~hard~negatives}}$ & 0.7140\\
    -- CT-M2-OneLook-SE$_{\text{MNR~w/~hard~negatives}}$ & 0.661\\
    -- CT-M2-BestLoss-SE$_{\text{MNR~w/~hard~negatives}}$ & 0.7032\\
    -- CT-M2-Complete-SE$_{\text{MNR~w/~hard~negatives}}$ & 0.7081\\
    -- CT-M3-OneLook-SE$_{\text{MNR~w/~hard~negatives}}$ & 0.6753\\
    -- CT-M3-BestLoss-SE$_{\text{MNR~w/~hard~negatives}}$ & 0.6845\\
    -- CT-M3-Complete-SE$_{\text{MNR~w/~hard~negatives}}$ & 0.6848\\
    \hline
     \textit{Complete QQP (102k training samples)} &\\
      -- CT-M1-Complete-SE$_{\text{MNR~w/~hard~negatives}}$   & 0.7394\\
    \hline
    \textit{Complete QQP and AllNLI (378k training samples)} &\\
      -- CT-M1-Complete-SE$_{\text{MNR~w/~hard~negatives}}$   & \textbf{0.7485}\\
    \hline
    \end{tabular}
    \caption{Overall performance of the evaluated sentence encoders across 18 datasets.}
    \label{average_results}
\end{table}

Furthermore, we investigated different pooling methods in our best-performing sentence encoder, CT-M1-Complete-SE$_{\text{MNR~w/~hard~negatives}}$. We considered four different strategies: (i) mean pooling with attention, (ii) [CLS] embedding, (iii) max-pooling, and (iv) mean pooling without attention. We utilized the 18 datasets listed in Table \ref{labelled_datasets} for this task. Results are summarized in Table \ref{pooling}. Across all datasets, mean pooling with attention achieved the highest scores (0.7485 on avg.). If attention is not considered while mean pooling, it seems to degrade the performance (0.5969 on avg.) and fall behind using [CLS] embedding (0.6612 on avg.). Max pooling seems to be the worst (0.3372 on avg.) among these four pooling strategies.

\begin{table}
\scriptsize
\centering
\begin{tabular}{lcccc}
\hline
Dataset & \multicolumn{1}{l}{\textbf{Mean (w/ atten.)}} & \multicolumn{1}{l}{\textbf{CLS}} & \multicolumn{1}{l}{\textbf{Max}} & \multicolumn{1}{l}{\textbf{Mean (w/o atten.)}} \\
\hline \hline
D-01                           & 0.7191                                 & 0.641                   & 0.3484                  & 0.4905                               \\
D-02                         & 0.8070                                  & 0.7048                  & 0.3114                  & 0.5461                               \\
D-03                          & 0.6128                                 & 0.5313                  & 0.2793                  & 0.4256                               \\
D-04                             & 0.6872                                 & 0.6082                  & 0.2810                   & 0.5481                               \\
D-05                               & 0.6744                                 & 0.5825                  & 0.2677                  & 0.5390                                \\
D-06                         & 0.6960                                  & 0.6010                   & 0.2821                  & 0.4769                               \\
D-07                        & 0.8742                                 & 0.7602                  & 0.2840                   & 0.7223                               \\
D-08                            & 0.7027                                 & 0.6420                   & 0.4780                   & 0.6061                               \\
D-09                           & 0.9021                                 & 0.8136                  & 0.6086                  & 0.7532                               \\
D-10                            & 0.7167                                 & 0.6449                  & 0.4864                  & 0.5757                               \\
D-11                              & 0.8982                                 & 0.7781                  & 0.2259                  & 0.7946                               \\
D-12                         & 0.8342                                 & 0.7414                  & 0.3948                  & 0.6506                               \\
D-13                   & 0.7874                                 & 0.6856                  & 0.3064                  & 0.6472                               \\
D-14                      & 0.7762                                 & 0.6799                  & 0.356                   & 0.6083                               \\
D-15                 & 0.7708                                 & 0.6816                  & 0.3635                  & 0.5831                               \\
D-16    & 0.6549                                 & 0.5921                  & 0.2875                  & 0.5865                               \\
D-17 & 0.5874                                 & 0.5331                  & 0.2226                  & 0.5406                               \\
D-18                             & 0.7722                                 & 0.6814                  & 0.2873                  & 0.6507                               \\ \hline
                                     & 0.7485                           & 0.6612            & 0.3372            & 0.5969           \\
\hline
\end{tabular}
\caption{Performance of different pooling methods in our best-performing sentence encoder.}
\label{pooling}
\end{table}

\subsection{Inference time analysis}

\begin{table}
\scriptsize
\centering

\begin{tabular}{lllllll}
\hline
& \multicolumn{6}{c}{Tokenization (in \textit{milliseconds})}\\ \hline
& mean   & std.   & min    & Q1     & Q2     & Q3       \\ \hline \hline
microsoft/mpnet-base              & 0.0796 & 0.0224 & 0.0467 & 0.0656 & 0.0755 & 0.0969 \\
vinai/bertweet-covid19-base-cased & 0.1868 & 0.0725 & 0.0818 & 0.1457 & 0.1823 & 0.2291 \\
bert-base-cased                   & 0.0772 & 0.0183 & 0.0459 & 0.0670 & 0.0743 & 0.0919 \\
roberta-base                      & 0.0743 & 0.0193 & 0.0420 & 0.0643 & 0.0679 & 0.0910  \\
xlm-roberta-base                  & 0.0854 & 0.0230 & 0.0471 & 0.0736 & 0.0847 & 0.1016 \\
albert-base-v2                    & 0.0966 & 0.0263 & 0.0579 & 0.0836 & 0.0883 & 0.1145  \\
google/electra-base-discriminator & 0.0875 & 0.0206 & 0.0591 & 0.0710 & 0.0851 & 0.1032 \\ \hline
Sentence-Transformers & 0.0791 & 0.0215 & 0.0461 & 0.0642 & 0.0773 & 0.0943 \\
SimCSE & 0.0744 & 0.0203 & 0.0411 & 0.0644 & 0.0699 & 0.0909 \\
\hline
CT-M1       & 0.0759 & 0.0202 & 0.0429 & 0.0662 & 0.0713 & 0.0930 \\
CT-M2       & 0.0800 & 0.0229 & 0.0486 & 0.0674 & 0.0754 & 0.0930  \\
CT-M3      & 0.1873 & 0.0711 & 0.0880 & 0.1487 & 0.1834 & 0.2268 \\
\hline
\end{tabular}

\vspace{1cm}

\begin{tabular}{lllllll}
\hline
& \multicolumn{6}{c}{Embeddings generation (in \textit{milliseconds})}           \\ \hline
& mean   & std.   & min    & Q1     & Q2     & Q3     \\ \hline \hline
microsoft/mpnet-base              & 0.0047 & 0.0039 & 0.0025 & 0.0032 & 0.0036 & 0.0045 \\
vinai/bertweet-covid19-base-cased & 0.0031 & 0.0046 & 0.0009 & 0.0011 & 0.0016 & 0.0026 \\
bert-base-cased                   & 0.0031 & 0.0044 & 0.0013 & 0.0013 & 0.0016 & 0.0025 \\
roberta-base                      & 0.0031 & 0.0045 & 0.0009 & 0.0013 & 0.0016 & 0.0025 \\
xlm-roberta-base                  & 0.0032 & 0.0047 & 0.0009 & 0.0011 & 0.0016 & 0.0026 \\
albert-base-v2                    & 0.0917 & 0.1302 & 0.0015 & 0.0024 & 0.0040 & 0.2190 \\
google/electra-base-discriminator & 0.0031 & 0.0042 & 0.0012 & 0.0014 & 0.0018 & 0.0026 \\ \hline
Sentence-Transformers             & 0.0046 & 0.0039 & 0.0024 & 0.0032 & 0.0036 & 0.0043 \\
SimCSE                            & 0.0031 & 0.0046 & 0.0009 & 0.0011 & 0.0015 & 0.0026 \\ \hline
CT-M1       & 0.0032 & 0.0046 & 0.0009 & 0.0011 & 0.0017 & 0.0029 \\
CT-M2       & 0.0031 & 0.0045 & 0.0009 & 0.0012 & 0.0016 & 0.0026 \\
CT-M3      & 0.0032 & 0.0047 & 0.0009 & 0.0011 & 0.0015 & 0.0025\\
\hline
\end{tabular}

\caption{Inference times of the baselines and CrisisTransformers.}
\label{complexity}
\end{table}

Given that each model utilizes a distinct set of tokens for tokenization and considering the varying number of parameters in each model, we conducted a comprehensive analysis of the inference time for both the baselines and CrisisTransformers, focusing on two key tasks: tokenization and embedding generation. Tokenization entails the creation of input identifiers and attention masks, while embedding generation encompasses feeding the outputs of tokenization into the model and producing mean token embeddings while considering the attention masks. These analyses were carried out using an Intel(R) Xeon(R) Gold 6326 CPU @ 2.90GHz alongside an 80GB A100 Nvidia GPU. We utilized native tokenizers from the latest release of each model. For every dataset listed in Table \ref{labelled_datasets}, we executed tokenization and embedding generation processes. We report the average processing times (measured in milliseconds) along with the standard deviation, minimum time, and Quartiles in Table \ref{complexity}.

In terms of tokenization, CT-M1 outperforms MPNet, BERTweet, BERT, XLM-RoBERTa, ALBERT, and Sentence-Transformers in average performance, despite having the largest vocabulary size. It also exhibits similar performance to RoBERTa and SimCSE. Among CrisisTransformers, CT-M3, which is based on BERTweet, demonstrates the slowest tokenization speed performance, comparable to BERTweet. Regarding embedding generation, CrisisTransformers exhibit nearly identical performance (0.0031--0.0032ms) to BERTweet, BERT, RoBERTa, XLM-RoBERTa, and SimCSE. MPNet, ALBERT, and Sentence-Transformers show much higher embedding generation times.

\subsection{Intended uses and limitations}
CrisisTransformers offers a selection of 8 pre-trained models alongside a sentence encoder. Similar to BERT and RoBERTa, the pre-trained models are designed for fine-tuning in downstream tasks (that require an entire sentence to make decisions), such as sequence classification and token classification. Additionally, the sentence encoder, similar to Sentence-Transformers, is intended for generating semantically meaningful sentence embeddings for tasks such as semantic search, clustering, and topic modeling.

The training corpus utilized by CrisisTransformers comprised a substantial volume of unfiltered tweets, inherently containing non-neutral content. Consequently, similar to RoBERTa and BERTweet, both the pre-trained models and their fine-tuned versions are prone to biased predictions. Biased predictions in this context refer to the tendency of the models to produce outputs that favour or exhibit partiality towards certain groups, perspectives, or sentiments present in the training data. Moreover, these models are specifically designed for processing crisis-related social media texts. Despite this focus, we observed that the CT-M2 and CT-M3 variants, which are built upon RoBERTa and BERTweet, respectively, also exhibit strong performance when applied to tweets from diverse domains (refer \ref{non-crisis-domains}). This efficacy can be attributed to the robustness of their original base models, supplemented by additional training data provided during this study. Furthermore, at this stage, CrisisTransformers can process only English-language tweets. As a part of future work, we plan to release their multi-lingual versions \cite{reimers2020making}.

\section{Conclusion}
\label{conclusion}

In this study, we introduced CrisisTransformers, an ensemble of pre-trained language models and sentence encoders designed for processing crisis-related social media texts. The pre-trained models were trained on a large-scale corpus of over 15 billion word tokens sourced from tweets associated with more than 30 crisis events that occurred between 2006 and 2023. Additionally, we fine-tuned the pre-trained models using siamese and triplet networks to create sentence encoders. Existing models and CrisisTransformers were evaluated on 18 crisis-specific datasets for classification and sentence encoding tasks. Our pre-trained models outperform strong baselines across all 18 datasets in classification tasks, and our best-performing sentence encoder improves the state-of-the-art by 17.43\% in sentence encoding tasks. We publicly release CrisisTransformers, which include 8 variants of pre-trained models and the best-performing sentence encoder, hoping that they will serve as a robust baseline for tasks that involve processing crisis-related social media texts.

CrisisTransformers offers checkpoints of models trained from scratch (CT-M1) and those initialized with RoBERTa's weights (CT-M2) and BERTweet's weights (CT-M3). During experimentations, we observed that pre-trained models (CT-M2 and CT-M3), which undergo further pre-training, leverage existing knowledge for efficient initial convergence, unlike randomly initialized CT-M1. CT-M2 and CT-M3 exhibited rapid initial drops in loss; CT-M3 later aligned with CT-M1 in terms of final loss. All models plateaued, implying convergence. In classification, CT-M1 performed best on 7 datasets, CT-M2 on 9, and CT-M3 on 2. Regarding sentence encoding, CT-M1 outperformed in 11 datasets, CT-M2 on 6, and CT-M3 on 1. Considering the training objectives, models trained with hard negatives achieved the highest scores across 11 datasets, which remains in line with what has been reported in the literature. We noticed that the CT-M1 at the lowest loss utilizing only 10k training samples with the MNR with hard negatives training objective outperformed the state-of-the-art Sentence-Transformers (trained on 1 billion samples) by a significant margin of 12\%. By increasing the training samples to 378k using the QQP+AllNLI datasets, the performance improved further to 17.43\%. This observation confirmed that domain-specific pre-trained models demonstrate significant improvements over general-purpose models in sentence encoding tasks. Going forward, our future objectives include training the sentence encoders on a scale similar to Sentence-Transformers. Also, the proposed models process only English-language tweets. As a future task, we aim to release their multi-lingual versions.

\section*{Acknowledgements} This study is supported by the Melbourne Research Scholarship from the University of Melbourne, Australia. This research was undertaken using the LIEF HPC-GPGPU Facility hosted at the University of Melbourne, which was established with the assistance of LIEF Grant LE170100200. The cloud infrastructure required to maintain \textit{COV19Tweets} over the last three years was provided by DigitalOcean. We appreciate the insights provided by Dat Quoc Nguyen (BERTweet's co-author) during the pre-training phase of CrisisTransformers.

\section*{CRediT authorship contribution statement}
\textbf{Rabindra Lamsal} performed Conceptualization, Data curation, Methodology, Software, Visualization, Writing--first draft. \textbf{Maria Rodriguez Read} and \textbf{Shanika Karunasekera} performed Conceptualization, Supervision, Writing--Review \& Editing.

\section*{Declaration of competing interest}
All authors declare that they have no known competing financial interests or personal relationships that could have appeared to influence the work reported in this paper.

 \bibliographystyle{elsarticle-num} 
 \bibliography{cas-refs}





\newpage

\appendix

\begin{landscape}
\section{Performance of baselines and CrisisTransformers on tweets from general domains}
\label{non-crisis-domains}

The table below summarizes the performance of baselines and CrisisTransformers on tweets from non-crisis domains. The same training configurations discussed in Section \ref{finetuning} were applied for fine-tuning. We report average F1-macro scores from the five runs.

\begin{table}[h]
\scriptsize
\begin{tabular}{lccccccccccc}
\hline
& Emoji \cite{barbieri2018semeval}                    & Emotion \cite{mohammad2018semeval} & Hate \cite{basile-etal-2019-semeval} & Irony \cite{van2018semeval} & Offensive \cite{zampieri2019semeval} & Sentiment \cite{rosenthal2017semeval} & Abortion$^*$ & Atheism$^*$ & Climate$^*$ & Feminist$^*$ & Hillary$^*$ \\ \hline \hline
MPNet              & 0.2542 & 0.7878                      & 0.4973                   & 0.7753                    & 0.7974                        & 0.6992                        & 0.5916                       & 0.6707                      & 0.5454                      & 0.622                        & 0.5898                   \\
BERTweet & 0.2939                     & 0.8038                      & 0.5391                   & 0.8277                    & 0.7917                        & 0.7133                        & 0.5738                       & 0.6904                      & 0.5613                      & 0.5923                       & 0.5736                    \\
BERT                   & 0.3043                     & 0.7583                      & 0.5                      & 0.6569                    & 0.7999                        & 0.6862                        & 0.5764                       & 0.598                       & 0.4763                      & 0.523                        & 0.5515                        \\
RoBERTa                      & 0.3230                     & 0.7854                      & 0.479                    & 0.4181                    & 0.7853                        & 0.7102                        & 0.5842                       & 0.7081                      & 0.5698                      & 0.6393                       & 0.6687                    \\
XLM-RoBERTa                  & 0.3177                     & 0.7676                      & 0.5013                   & 0.7009                    & 0.7825                        & 0.7007                        & 0.5973                       & 0.6871                      & 0.5339                      & 0.5598                       & 0.5736                        \\
ALBERT                    & 0.2360                     & 0.7119                      & 0.5125                   & 0.6873                    & 0.7939                        & 0.6846                        & 0.4914                       & 0.5885                      & 0.4891                      & 0.442                        & 0.4447                         \\
Electra & 0.2414                     & 0.7793                      & 0.4781                   & 0.7088                    & 0.8081                        & 0.7115                        & 0.5575                       & 0.5927                      & 0.5134                      & 0.5256                       & 0.5184                              \\ \hline
CT-M1-BestLoss                    & 0.3279                     & 0.7808                      & 0.4962                   & 0.7457                    & 0.8134                        & 0.7112                        & 0.6                          & 0.6694                      & 0.5719                      & 0.5845                       & 0.6527                    \\
CT-M1-Complete                    & 0.3294                     & 0.7869                      & 0.5077                   & 0.6175                    & 0.8069                        & 0.7214                        & 0.5856                       & 0.6776                      & 0.5641                      & 0.5951                       & 0.6415                 \\
CT-M2-OneLook                     & 0.3323                     & 0.794                       & 0.5002                   & 0.5422                    & 0.8112                        & 0.7031                        & 0.6179                       & 0.679                       & 0.5555                      & 0.6074                       & 0.6155               \\
CT-M2-BestLoss                    & 0.3323                     & 0.7841                      & 0.4844                   & 0.5337                    & 0.8107                        & 0.7133                        & 0.592                        & 0.688                       & 0.5522                      & 0.6076                       & 0.6526                  \\
CT-M2-Complete                    & 0.3333                     & 0.7856                      & 0.4896                   & 0.5554                    & 0.8068                        & 0.7071                        & 0.5946                       & 0.6949                      & 0.5545                      & 0.6014                       & 0.6432                 \\
CT-M3-OneLook                     & 0.2854                     & 0.7729                      & 0.5512                   & 0.8267                    & 0.8009                        & 0.7147                        & 0.626                        & 0.6593                      & 0.5727                      & 0.5954                       & 0.6416                 \\
CT-M3-BestLoss                    & 0.2975                     & 0.7838                      & 0.5223                   & 0.8256                    & 0.8142                        & 0.7215                        & 0.6272                       & 0.6154                      & 0.562                       & 0.6249                       & 0.6388               \\
CT-M3-Complete                    & 0.2940                     & 0.7904                      & 0.5362                   & 0.8157                    & 0.8103                        & 0.7239                        & 0.624                        & 0.5827                      & 0.5643                      & 0.6127                       & 0.652      \\
\hline
\end{tabular}
$^*$stance detection \cite{mohammad2016semeval}
\end{table}
    \end{landscape}

\section{Configurations of CrisisTransformers}
\label{config}
\begin{table}[h!]
    \centering
    \scriptsize
    \begin{tabular}{c|c|c|c}
    \hline
         & CT-M1 & CT-M2 & CT-M3 \\
         \hline \hline
architecture   & RobertaForMaskedLM & RobertaForMaskedLM & RobertaForMaskedLM \\
attention\_probs\_dropout\_prob & 0.1 & 0.1 & 0.1 \\
hidden\_act & gelu & gelu & gelu \\
hidden\_dropout\_prob & 0.1 & 0.1 & 0.1 \\
hidden\_size & 768 & 768 & 768 \\
intermediate\_size & 3072 & 3072 & 3072 \\
layer\_norm\_eps & 1e-12 & 1e-05 & 1e-05 \\
max\_position\_embeddings & 130 & 514 & 130 \\
num\_attention\_heads & 12 & 12 & 12 \\
num\_hidden\_layers & 12 & 12 & 12 \\     
vocab\_size & 64,000 & 50,265 & 64001 \\
      
      \hline
    \end{tabular}
\end{table}

\end{document}